\RequirePackage{fix-cm}
\documentclass[smallcondensed]{svjour3}     
\smartqed  

\usepackage{booktabs}       
\usepackage{subfig}
\usepackage{multirow}
\usepackage{multicol}
\usepackage{array}
\usepackage{amsmath}
\usepackage{amsfonts}
\usepackage{bm}
\usepackage{graphicx}
\usepackage{algorithm}
\usepackage{algorithmicx}
\usepackage{algpseudocode}
\usepackage[switch]{lineno}

\usepackage{import}
\usepackage{xifthen}
\usepackage{pdfpages}
\usepackage{transparent}
\usepackage{url}
\usepackage{color}
\usepackage{epstopdf}

\newcommand{\ca}[1]{\bm{\mathcal{#1}}}

\newtheorem{hypothesis}{Hypothesis}

\floatname{algorithm}{Algorithm}

\renewcommand{\thefootnote}{\arabic{footnote}}
\begin{document}

\title{Heuristic Rank Selection with Progressively Searching Tensor Ring Network%
}

\author{Nannan~Li~\textsuperscript{$\dag$}\and
        Yu~Pan~\textsuperscript{$\dag$}\and
        Yaran~Chen~\textsuperscript{$\ast$}  \and
        Zixiang~Ding \and
        Dongbin~Zhao \and
        Zenglin~Xu~\textsuperscript{$\ast$} 
}


\institute{ Nannan Li, Yaran Chen, Zixiang ding and Dongbin Zhao \at 
            State Key Laboratory of Management and Control for Complex Systems,
            Institute of Automation, Chinese Academy of Sciences, Beijing 100190, China\\
            School of Artificial Intelligence, University of Chinese Academy of Sciences, Beijing 101408, China \\
            \email{$\{linannan2017, chenyaran2013, Dingzixiang2018, dongbin.zhao\}$@ia.ac.cn}           
            \and
            Yu Pan \at
            Harbin Institute of Technology(Shenzhen), Shenzhen, China\\
            \email{$iperryuu$@gmail.com}
            \and
            Zenglin~Xu \at
            Harbin Institute of Technology(Shenzhen), Shenzhen, China\\
            Pengcheng Lab, Shenzhen, China\\
            \email{$zenglin$@gmail.com}
}

\date{Received: date / Accepted: date}

\maketitle

\def\thefootnote{$\dag$}\footnotetext{These authors contributed equally to this work.}
\def\thefootnote{$\ast$}\footnotetext{These authors are corresponding authors of this work.}

\renewcommand{\thefootnote}{\arabic{footnote}}

\begin{abstract}

Recently, Tensor Ring Networks (TRNs) have been applied in deep networks, achieving remarkable successes in compression ratio and accuracy.
Although highly related to the performance of TRNs, rank selection is seldom studied in 
previous works and usually set to equal in experiments.
Meanwhile, there is not any heuristic method to choose the rank, and an enumerating way to find appropriate rank is extremely time-consuming.
Interestingly, we  
discover that part of 
the rank elements is sensitive and usually aggregate in a narrow region, namely an interest region.
Therefore, based on the above phenomenon, we propose a novel progressive genetic algorithm named Progressively Searching Tensor Ring Network Search (PSTRN), which has the ability to find optimal rank precisely and efficiently.
Through the evolutionary phase and progressive phase, PSTRN can converge to the interest region quickly and harvest good performance.
Experimental results show that PSTRN can significantly reduce the complexity of seeking rank, compared with the enumerating method. Furthermore, our method is
validated  
on public benchmarks like
MNIST,
CIFAR10/100, UCF11 and HMDB51, achieving the state-of-the-art performance. 

\end{abstract}

\section{Introduction}
Deep neural networks have made great successes in various areas, such as image classification~\cite{DBLP:conf/cvpr/HeZRS16,DBLP:journals/corr/SimonyanZ14a,DBLP:conf/ijcnn/0003CDZ20}, autonomous driving \cite{chen2018multi,DBLP:journals/tamd/ZhaoCL17,DBLP:conf/ijcnn/LiZCZ18}, game artificial intelligence~\cite{DBLP:conf/cig/ShaoZLZ18,DBLP:conf/ssci/LiZZ19a} and so on~\cite{DBLP:conf/icml/LiS20,DBLP:conf/ijcnn/WangZPXX19,DBLP:conf/iconip/WangSL0ZX20}. 
However, parameters redundancy leads to two major drawbacks for deep neural networks: 1) difficult training, and 2) poor ability to run on resource-constrained devices (e.g., mobile phones~\cite{DBLP:journals/corr/KimPYCYS15} and Internet of Things (IoT) devices~\cite{DBLP:conf/sensys/LaneBGFK15}). To address these problems, Tensor Ring (TR) has been introduced to deep neural networks. With a ring-like structure as shown in Fig.~\ref{fig:TR}, TR can significantly reduce the parameters of Convolutional Neural Network (CNN)~\cite{DBLP:conf/cvpr/WangSEWA18} and Recurrent Neural Network (RNN)~\cite{DBLP:conf/aaai/PanXWYWBX19}, and even can achieve better results than uncompressed models in some tasks. Thus tensor ring is increasingly being researched.

However, as the crucial component of tensor ring, setting of rank (e.g. $R_{0} \sim R_{3}$ in Fig.~\ref{fig:TR}) is seldom investigated. In most of the existing works, it merely sets to be equal in whole network~\cite{DBLP:conf/cvpr/WangSEWA18}. Such an equal setting requires multiple manual attempts for a feasible rank value and often leads to a weak result. Fortunately, as shown in our synthetic experiment, we discover the relationship between the rank distribution and its performance. Experimental results demonstrate the link that part of rank elements with good performance will gather to the interest region. Then we extend this phenomenon to build our Hypothesis~\ref{hyp}. Utilizing the hypothesis, we design a heuristic algorithm to explore the potential power of tensor ring.

\begin{figure*}[t]
\centering
\includegraphics[width=1\columnwidth]{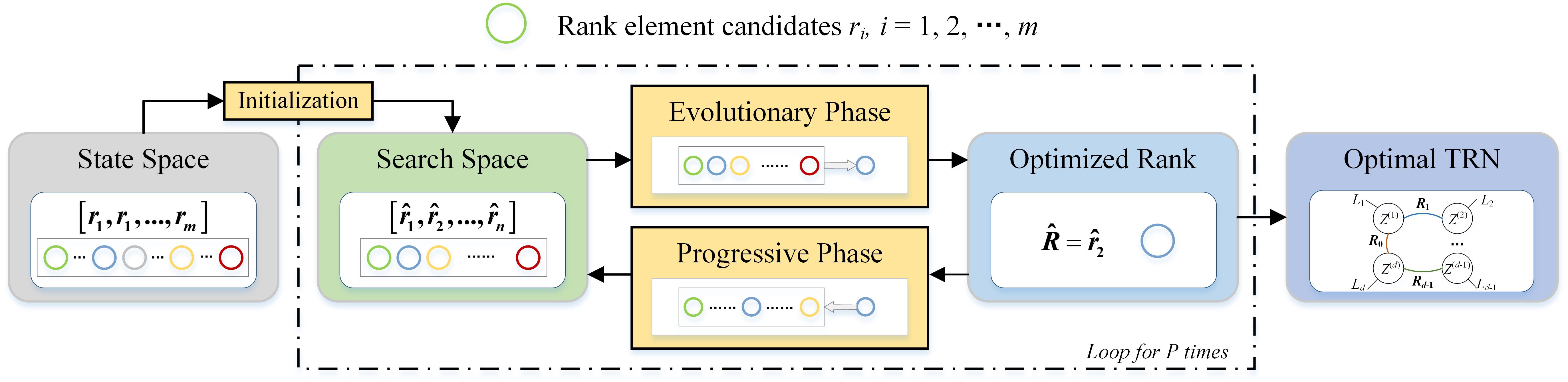} 
\caption{The overview of Progressive Searching Tensor Ring Network (PSTRN), where different color represents different rank element candidate. PSTRN initializes search space by sampling from state space, then alternately executes evolutionary phase and progressive phase for $P$ times to derive optimal TRN.}
\label{fig:framwork}
\end{figure*}

Specifically, we propose Progressive Searching Tensor Ring Network (PSTRN) inspired by Neural Architecture Search (NAS)~\cite{DBLP:conf/iclr/ZophL17}. Similarly, our approach is divided into three parts, 
\begin{itemize}
    \item search space: combinations of rank element candidates for TRN in evolutionary phase; 
    \item search strategy: the Non-dominated Sorted Genetic
Algorithm-II (NSGA-II)~\cite{DBLP:journals/tec/DebAPM02} to search rank;
    \item performance estimation strategy: stochastic gradient descent to train TRN.
\end{itemize}
The overall framework of PSTRN is illustrated in Fig.~\ref{fig:framwork}. 
In the searching process, we initialize search space first. Then through evolutionary phase, we derive optimized rank within search space.
Next, in order to draw near interest region, the proposed approach shrinks the bound of search space to the around of optimized rank during progressive phase. By alternately executing evolutionary phase and progressive phase, our algorithm can find rank with high performance.     
Additionally, on large-scale models (i.e. ResNet20/32~\cite{DBLP:conf/cvpr/HeZRS16} and WideResNet28-10~\cite{DBLP:conf/bmvc/ZagoruykoK16}), the performance estimation is time-consuming, which is harmful to search speed. So we employ a weight inheritance strategy~\cite{DBLP:conf/icml/RealMSSSTLK17} to accelerate the evaluation of rank.

Experimental results prove that PSTRN can obtain optimal rank of TRN according to the Hypothesis~\ref{hyp}. And our algorithm can compress LeNet5~\cite{lecun1998gradient} with compression ratio as 16x and 0.49\% error rate in MNIST~\cite{DBLP:journals/spm/Deng12} image classification task.
In TR-ResNets, our approach can achieve state-of-the-art performance on CIFAR10 and CIFAR100~\cite{krizhevsky2009learning}. PSTRN also exceeds TR-LSTM models that set rank elements equal on HMDB51 and UCF11. Furthermore, compared with the enumerating method, our work can greatly reduce the complexity of seeking rank. 
Overall, our contributions can be summarized as follows:
\begin{enumerate}
\item 
PSTRN can search rank automatically instead of manual setting. 
At the meantime, 
The time cost is reduced significantly by progressively searching, compared with an enumerating method.
\item To speed up the search on large-scale model, our proposed method adopts weight inheritance into the search process. And the proposed method achieves about $200 \times$ speed-up ratio on classification tasks of CIFAR10/100 datasets.
\item As a heuristic approach based on the Hypothesis~\ref{hyp}, our algorithm can achieve better performance with fewer parameters than existing works. All the experimental results demonstrate the rationality of the Hypothesis that is first found by us.
\end{enumerate}

\section{Background}
In this section, we will introduce the tensor background and some related works that consist of rank fixed method and rank selection method. The rank fixed method is the work that sets rank manually, while rank selection method means the work of learning the rank.

\subsection{Tensor Background}  
In this part, we would like to introduce the background of tensor.

\begin{figure}[t]
\centering
\subfloat[A Tensor]{
    \includegraphics[width=0.18\columnwidth
]{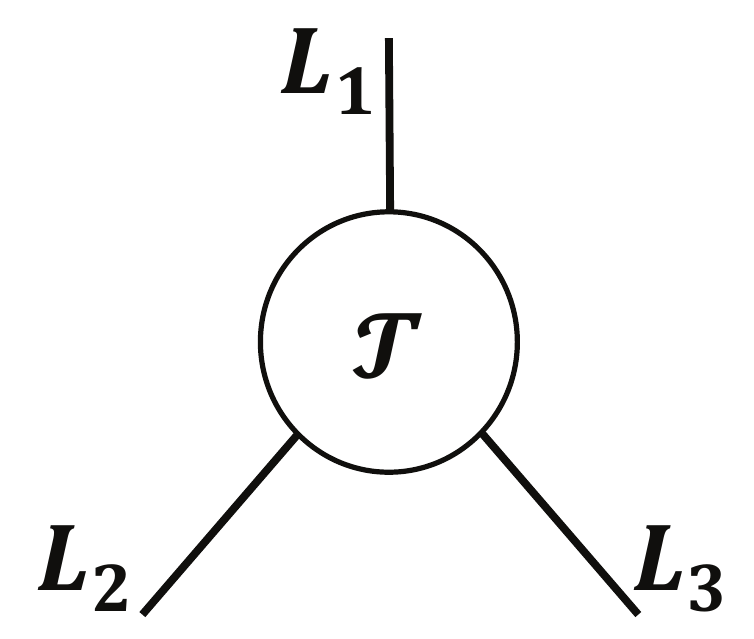}
    \label{fig:tensor}
}~~~~~
\subfloat[Tensor Contraction]{
    \includegraphics[width=0.3\columnwidth
]{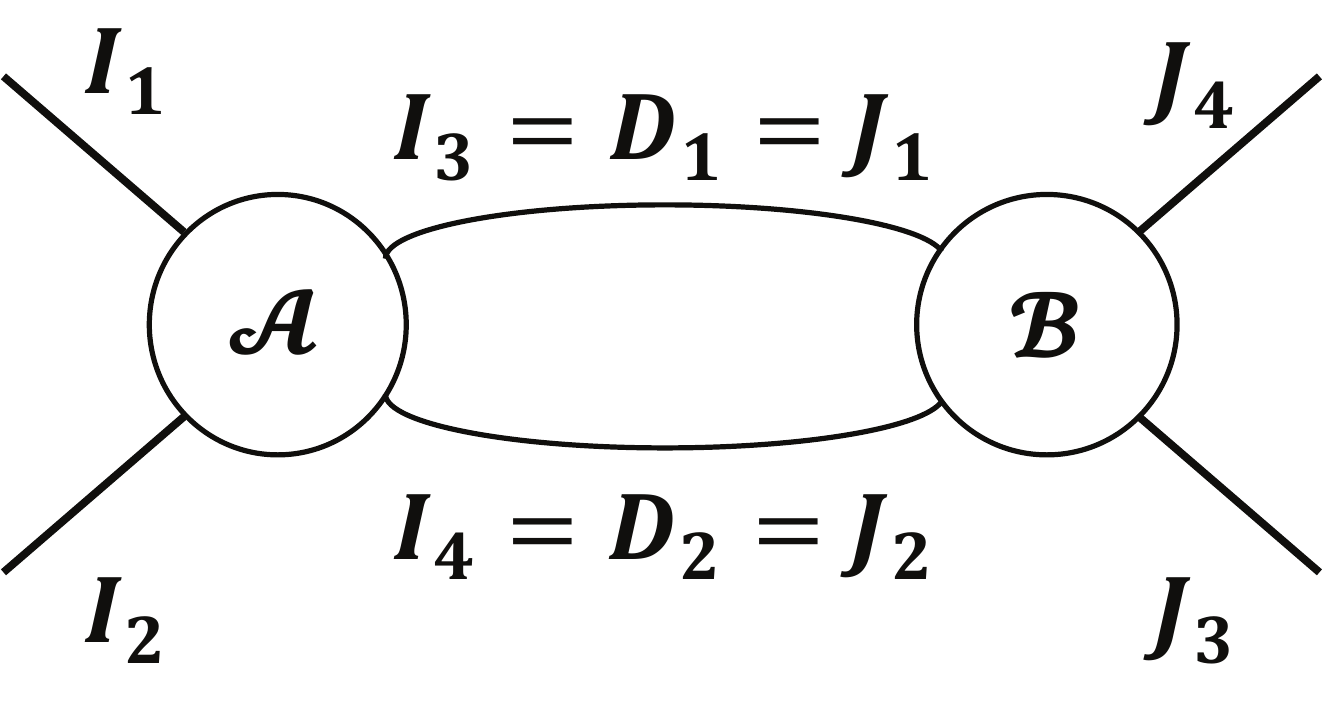}
    \label{fig:contracrion}
}
\caption{Tensor diagrams. \protect\subref{fig:tensor} presents the graphical notation of a tensor $\ca{T} \in \mathbb{R}^{L_1\times L_2 \times L_3}$. \protect\subref{fig:contracrion} demonstrates the contraction between two 4-order tensors, which is the contraction between $\ca{A}$ and $\ca{B}$.}
\label{fig:tensor_notation}
\end{figure}

\subsubsection{Notation}
A tensor is a high-order array. In this paper, a $d$-order tensor $\ca{T} \in \mathbb{R}^{L_1\times L_2 \dots \times L_d} $ is denoted by a boldface Euler script letter. With all subscripts fixed, each element of a tensor is expressed as: $\ca{T}_{l_1,l_2,\dots l_d}\in \mathbb{R}$. Given a subset of subscripts, we can get a sub-tensor. For example, given a subset $\{L_1=l_1, L_2=l_2\}$, we can obtain a sub-tensor $\ca{T}_{l_1, l_2} \in \mathbb{R}^{L_3 \dots \times L_d}$.  Fig.~\ref{fig:tensor_notation} draws
the tensor diagrams that present the graphical notations and the essential operations.

\subsubsection{Tensor Contraction} Tensor contraction can be performed between two tensors if some of their dimensions are matched. As shown in Fig.~\ref{fig:contracrion}, given two  4-order tensors $\ca{A}\in\mathbb{R}^{I_1\times I_2\times I_3\times I_4}$ and $\ca{B} \in \mathbb{R}^{J_1\times J_2 \times J_3 \times J_4}$, when $I_3 = D_1 = J_1$ and $I_4 = D_2 = J_2$, the contraction between these two tensors results in a tensor with the size of  $I_1\times I_2 \times J_3 \times J_4$, where the matching dimension is reduced, as shown in  equation:
\begin{equation}
    (\ca{A}\ca{B})_{i_1, i_2, j_3, j_4} 
= \sum_{m=1}^{D_1}\sum_{n=1}^{D_2} \ca{A}_{i_1, i_2, m, n }\ca{B}_{m, n, j_3, j_4}.
\label{eq:tensor_contraction}
\end{equation}

\subsubsection{Tensorization}
Given a matrix $\mathbf{M} \in \mathbb{R}^{ I \times O}$, we transfer it into a new tensor 
$$\ca{C} \in \mathbb{R}^{I_0 \times I_1 \times \dots \times I_{M-1} \times O_0 \times O_1 \times \dots \times O_{N-1}}$$
satisfying the equation:
\begin{equation}
    \prod_{i=1}^{M}I_i = I, ~~~\prod_{j=1}^{N}O_j = O,
\end{equation}
where $M, N$ are the number of the input nodes and output nodes respectively. Therefore, a corresponding element of $\mathbf{W}_{i, o}$ is $\ca{C}_{i_0, \dots, i_{m-1}, o_0, \dots, o_{n-1}}$, where $i \in \{1, \dots, I\}$, $o \in \{1, \dots, O\}$, $i_\ast \in \{1, \dots, I_\ast\}$, $o_\ast \in \{1, \dots, O_\ast\}$ are indexes, following the rule~\footnote{Here, define $\prod^{0}_{v}{(\bullet)} = 1, v > 0$.}
\begin{equation}
    i = \sum^{M}_{u=1}{\prod^{i_u-1}_{v=1}{I_v}(i_u-1)}, 
    o = \sum^{N}_{u=1}{\prod^{o_u-1}_{v=1}{O_v}(o_u-1)}.
\label{eq:w_x_y_shape}
\end{equation}

\subsubsection{Tensor Ring Format (TRF)}

\begin{figure}[t]
\centering
\includegraphics[width=0.75\columnwidth]{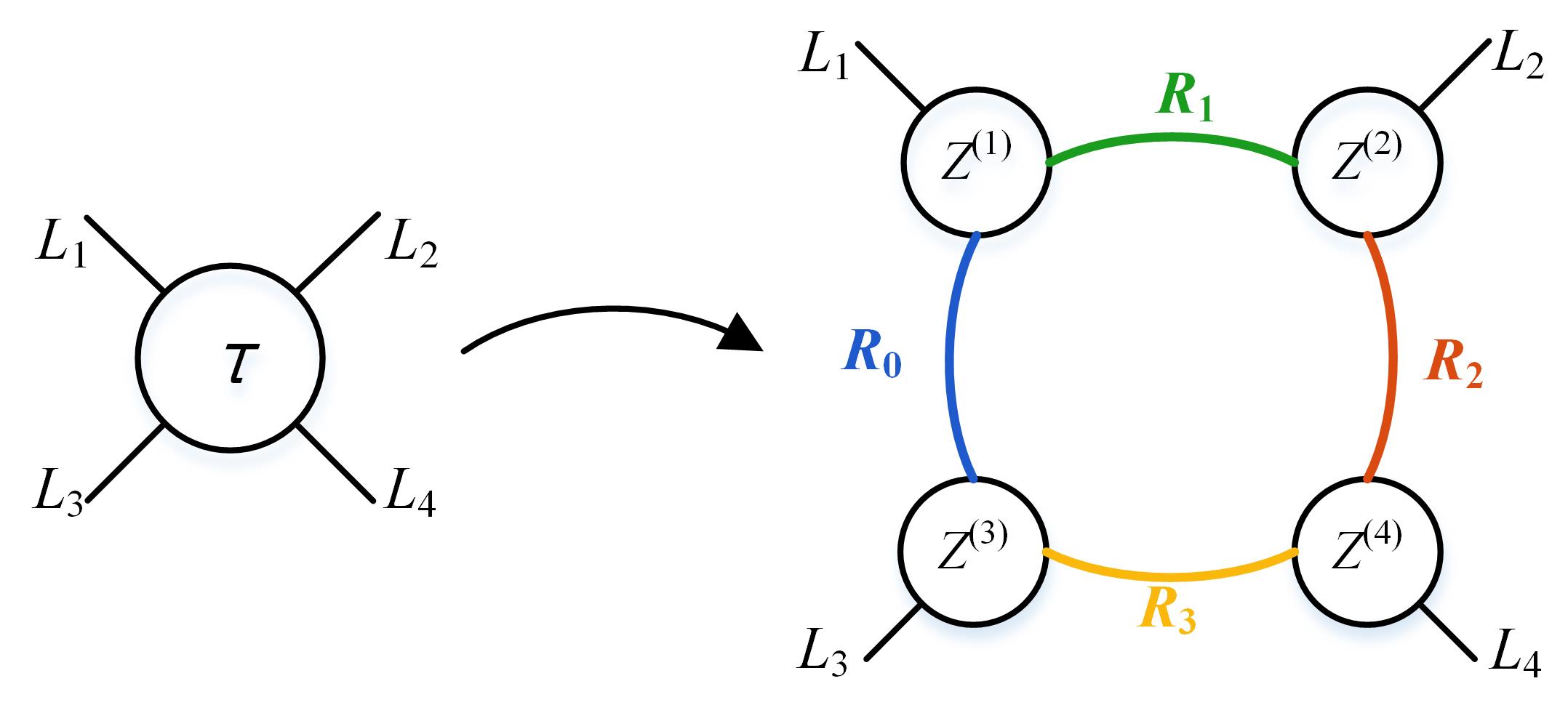}
\caption{The representations of TRF.}
\label{fig:TR}
\end{figure}
TRF is constructed with a series of $3$-order nodes linked one by one, forming a ring-like structure. The TRF of a $d$-order tensor can be formulated as
\begin{equation}
    \ca{T}_{l_1,l_2,\dots, l_d} 
    {=} \sum^{R_0, R_1, \dots, R_{d-1}}_{r_0, r_1, \dots, r_{d-1}}\notag \\
    \pmb{\mathcal{Z}}^{(1)}_{r_0,l_1, r_1}\pmb{\mathcal{Z}}^{(2)}_{r_1, l_2, r_2}\dots\pmb{\mathcal{Z}}^{(d)}_{r_{d-1}, l_d,r_0},
    \label{eq:TR_decom}
\end{equation}
where ${\rm\mathbf{R}} = \{R_i | i \in \{0, 1, \dots, d-1\}\}$ denotes the rank of TRF, and the symbol $\ca{Z}$ represents the tensor ring node. Fig. \ref{fig:TR} shows a graph structure of a simple TRF. 

Through replacing layers(e.g. convolutional layer, fully-connected layer) of a network with TRF, we can derive a TRN. 
                
\subsection{Rank Fixed}
Tensor ring decomposition has been successfully applied to the compression of deep neural networks. Wenqi et al.~\cite{DBLP:conf/cvpr/WangSEWA18} compress both the fully connected layers and
the convolutional layers of CNN with the equal rank elements for whole network. Yu et al.~\cite{DBLP:conf/aaai/PanXWYWBX19} replace the over-parametric input-to-hidden layer of LSTM with TRF, when dealing with high-dimensional input data. 
Rank of these models are determined via multiple manual attempts by manipulation, which requires much time.

\subsection{Rank Selection}
In this part, we would like to introduce the works of rank selection. Yerlan et al.~\cite{DBLP:conf/cvpr/IdelbayevC20} formulate the low-rank compression problem as a mixed discrete-continuous
optimization jointly over the rank elements and over the matrix elements. Zhiyu et al.~\cite{DBLP:conf/icassp/ChengLFB20} propose a novel rank selection scheme for tensor ring, which apply deep deterministic policy gradient to control the selection of rank. 
Their algorithms calculate the optimal rank directly from the trained weight matrix without the analysis of rank.
Different from them,
our approach is inspired by the relevance between the rank distribution and performance of Hypothesis~\ref{hyp}
, towards a better result.

\section{Methodology}
To verify the optimization of PSTRN on TRN, we choose two most commonly used deep neural networks for evaluation, i.e. Tensor Ring CNN (TR-CNN) and Tensor Ring LSTM (TR-LSTM).

In the section, we first present preliminaries of TR-CNN and TR-LSTM, including graphical illustrations of the two TR-based models. Then we elaborate on evolutionary phase and progressive phase of PSTRN. The implementation of weight inheritance will be given in final.

\subsection{Preliminaries}
\begin{figure}[t]
\centering
	\subfloat[TR-CNN]{
	    \includegraphics[width=0.35\columnwidth]{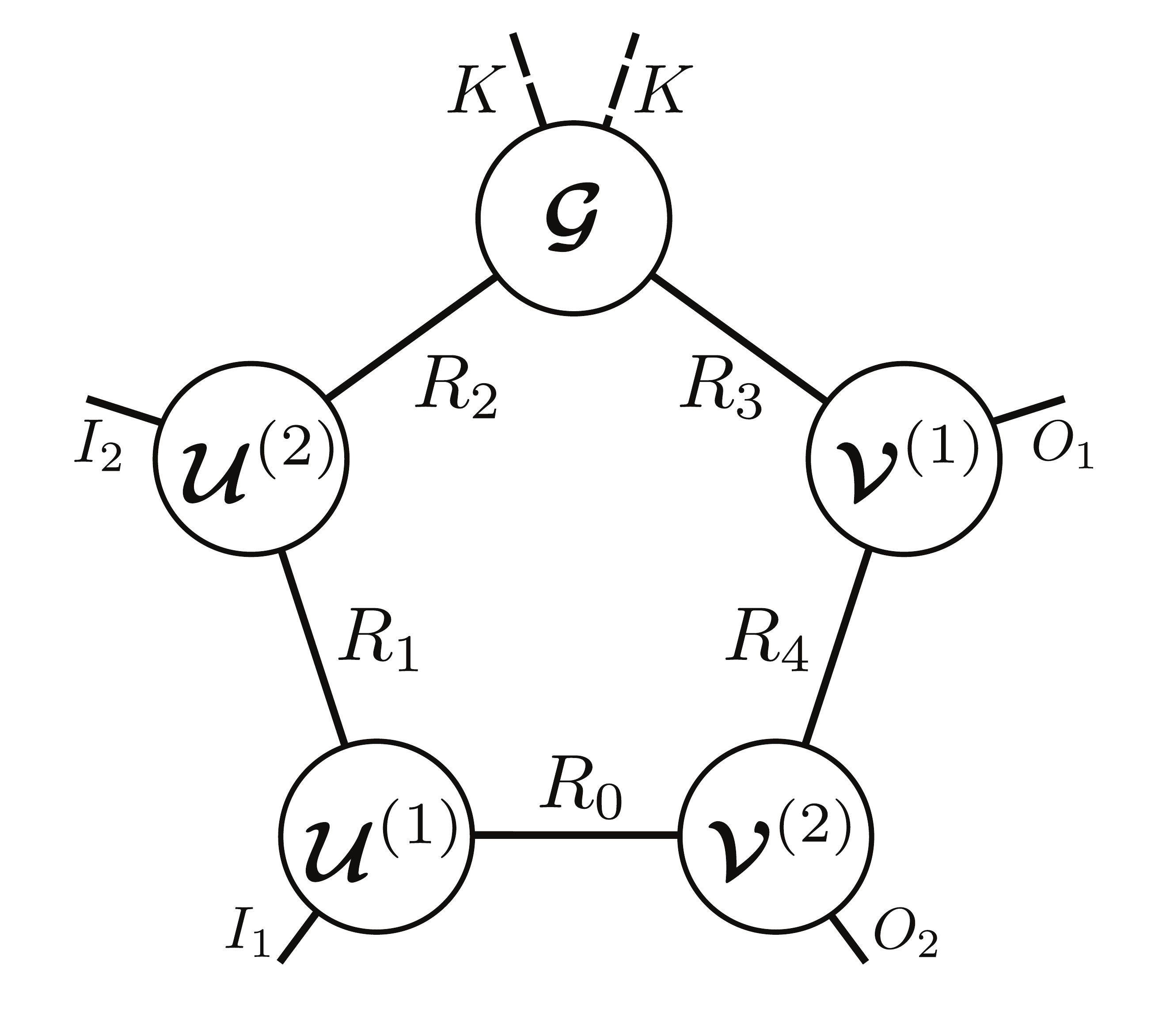}
	    \label{fig:tr_cnn}
	}
	\subfloat[TR-LSTM]{
	    \includegraphics[width=0.35\columnwidth]{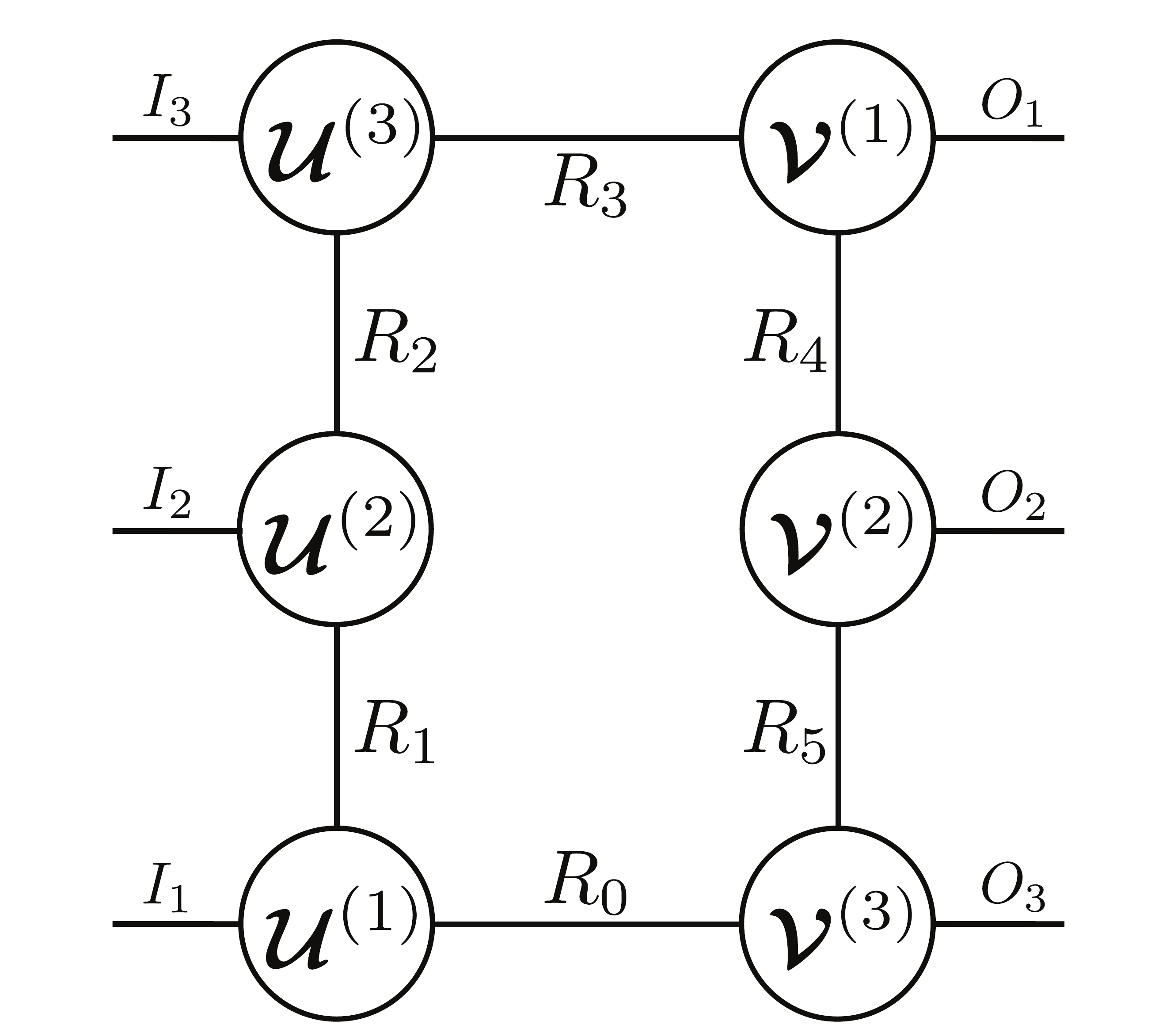}
	    \label{fig:tr_rnn}
	}
\caption{Tensor Ring Model}
\label{fig:tr_model}
\end{figure}
\subsubsection{TR-CNN}
For a convolutional core $\ca{C} \in \mathbb{R}^{K \times K \times C_{in} \times C_{out}}$ where $K$ denotes the kernel size, $C_{in}$ means the input channel and $C_{out}$ represents the output channel. We first reshape it as $\hat{\ca{C}} \in \mathbb{R}^{K \times K \times I_1 \times \dots \times I_{\alpha} \times O_1 \times \dots \times O_{\beta}}$, satisfying the rule
\begin{equation}
    C_{in} = \prod^{\alpha}_{i=1}{I_i},~~~C_{out} = \prod^{\beta}_{j=1}{O_j}.
\label{eq:cnn_channel}
\end{equation}

Then we decompose it into input nodes $\ca{U}^{(i)} \in \mathbb{R}^{{R_{i-1}}\times{I_{i}}\times{R_{i}}}, i \in \{1, 2, \dots, {\alpha}\}$, output nodes $\ca{V}^{(j)} \in \mathbb{R}^{{R_{{\alpha}+j}}\times{O_{i}}\times{R_{{\alpha}+j+1}}}, j \in \{1, 2, \dots, {\beta}\}$ and one convolutional node $\ca{G} \in \mathbb{R}^{K\times K\times{R_{\alpha}}\times{R_{{\alpha}+1}}}$, where $R_{{\alpha}+{\beta}+1} = R_0$.
An instance (${\alpha}=2$, ${\beta}=2$) is illustrated in Fig.~\ref{fig:tr_cnn}.
And the compression ratio of TR-CNN is calculated as
\begin{equation}
    C_{CNN} = \frac{K^2C_{in}C_{out}}
{
\sum^{\alpha}_{i=1}{R^2{I_{i}}} +
\sum^{\beta}_{j=1}{R^2{O_{j}}} +
K^2R^2
},
\label{eq:compression_ratio_cnn}
\end{equation}
where $R$ is a simplification of rank element. The TR-CNN is proposed by Wenqi et al.~\cite{DBLP:conf/cvpr/WangSEWA18}.

\subsubsection{TR-LSTM}

By replacing each matrix of the affine matrices $\mathbf{W}_{\ast} \in \mathbb{R}^{I \times O}$ of input vector $x \in \mathbb{R}^I$ with TRF in LSTM, we implement the TR-LSTM model as introduced by Yu et al.~\cite{DBLP:conf/aaai/PanXWYWBX19}.
Similar to TR-CNN, the nodes are combined by input nodes $\ca{U}^{(i)}$ and output nodes $\ca{V}^{(j)}$, and the decomposition needs to follow
\begin{equation}
      I = \prod^{\alpha}_{i=1}{I_i},~~~O = \prod^{\beta}_{j=1}{O_j}.
\label{eq:rnn_matrix}
\end{equation}

 A 6-node example is shown in Fig.~\ref{fig:tr_rnn}. Compression ratio of TR-LSTM can be computed as
\begin{equation}
    C_{RNN} = \frac{IO}{\sum^{\alpha}_{i=1}{R^2I_i}+\sum^{\beta}_{j=1}{R^2O_j}}.
\label{eq:compression_ratio_rnn}
\end{equation}

\begin{figure*}[t]
\centering
\includegraphics[width=1\columnwidth]{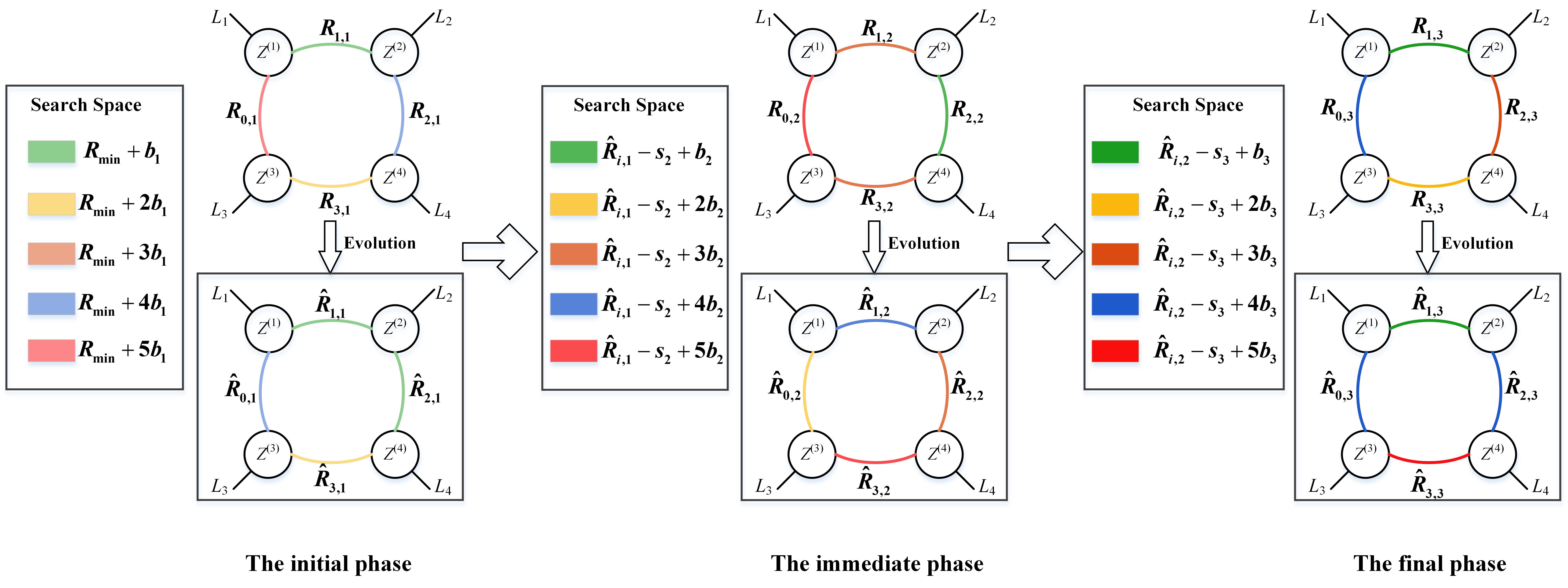} 
\caption{The overall pipeline of progressive phase, where different color represents different rank. The search space is sampled at interval $b$ within given range first. Then we gradually reduce the interval $b$. Obviously, progressive phase narrows
search space close to interest region progressively.}
\label{fig:progressive_evolve}
\end{figure*}

\subsection{Progressive Searching Tensor Ring Network}

In our search process, the rank ${\rm\mathbf{R}}$ of a TRN is formulated as
\begin{equation}
    {\rm\mathbf{R}} = \{R_{0}, R_{1}, \dots, R_{d-1} | R_{\ast} \in \{r_1, r_2, \dots, r_{m}\}\},
\label{eq:r}
\end{equation}
where $d$ is the number of rank elements, $r_{\ast}$ is a rank element candidate
, and $m$ is the quantity of rank element candidates. 
Full combinations of the rank elements (i.e. state space) can be calculated as
\begin{equation}
    S_{state} = {m}^d.
\end{equation}

Next, we would like to introduce the Hypothesis~\ref{hyp}, the extension of the aforementioned gathering phenomenon.
\begin{hypothesis}
When a shape-fixed TRN performs well, part or all of its rank elements are sensitive and each of them will tend to aggregate in a narrow region, which is called interest region.
\label{hyp}
\end{hypothesis}

According to Hypothesis~\ref{hyp}, the optimal rank can be found in the interest region.
It is a more efficient and accurate way to find a optimal rank in interest region rather than a much wider range of the whole rank element candidates.
Thus, we build an pipeline of PSTRN to achieve the purpose by two alternative procedures:
\begin{itemize}
    \item Evolutionary phase: finding good models in the search space and locating the interest region through well-performed models. 
    \item Progressive phase: calculating the width of a rough approximation of interest region and defining search space within this region.
\end{itemize}
Through these two procedures, the rank of a TRN will approach the interest region and be optimized.
Additionally, we apply weight inheritance to accelerate the training process. The pseudocode of the algorithm is shown as below, where $P$ is the number of progressive phase, and $G$ is the generations of each evolutionary phase.

\begin{algorithm}
\caption{Progressive Searching Tensor Ring Network}
\label{alg1}
\begin{algorithmic}
\Require Datasets $\mathcal{D}$, generations of evolutionary phase $\mathcal{G}$, 
\Statex $\qquad$ number of progressive phase $\mathcal{P}$.
\State Initialize the search space
\For{$p = 1,2, ... , P$}
    \If {Large-scale model (TR-ResNets)}
        \State Do warm-up of weights
    \EndIf
    \State Generate a set of randomized ranks following Eq.~(\ref{eq:r})
    \State Compute their accuracy and compression ratios via Eq.~(\ref{eq:compression_ratio_cnn}) and Eq.~(\ref{eq:compression_ratio_rnn})
    \For{$g = 1,2, ... , G$}
        \State Do selection, mutation and crossover
        \State Obtain the promising rank
    \EndFor
    \State Determine the search space for next phase by Eq.~(\ref{eq:space})
\EndFor
\Ensure A set of well-performed tensor ring rank elements
\end{algorithmic}
\end{algorithm}

\subsubsection{Evolutionary Phase}
As described in Hypothesis~\ref{hyp} that well-performed models aggregate in interest region, good models keep a high probability of appearing in interest region. Therefore, we determine interest region around the models with high performance. 

In PSTRN, we adopt multi-objectives  genetic algorithm NSGA-II~\cite{DBLP:journals/tec/DebAPM02} to search for TR-based models with high performance and few parameters.

A typical genetic algorithm requires two prerequisites, the genetic representation of solution domain (i.e. search space), and the fitness functions (i.e. classification accuracy and compression ratio) that is used to evaluate each individual. In the process,
an individual means the rank ${\rm\mathbf{R}}$ and each rank element $R_{\ast}$ is in $\{\hat{r}_1, \hat{r}_2, \dots, \hat{r}_n\}$ that is sampled from the whole rank element candidates.
The search space is a sub-space of the state space and can be calculated as 
\begin{equation}
    S_{search} = n^d.
\end{equation}

The method of choosing the search space will be introduced in the progressive phase. Classification accuracy is obtained by testing the model on the test dataset. And compression ratio of TR-CNN and TR-LSTM are calculated by Eq.~(\ref{eq:compression_ratio_cnn}) and Eq.~(\ref{eq:compression_ratio_rnn}) respectively. 

The key idea of the genetic algorithm is to evolve individuals via some genetic operations. 
At each evolutionary generation, 
the selection process preserves strong individuals
as a population and then sorts them according to their fitness function,
while eliminating weak ones. The retained strong individuals reproduce new children through mutation and crossover with a certain probability. After this, we obtain the new population consists of the new children and the retained strong individuals. 
The new population executes the evolution to derive next generation. When the termination condition is met, evolutionary phase stops and optimization of the rank will be completed. Eventually, taking the top-$k$ individuals into consider, we derive the most promising rank element $\hat{R}_{\ast}$ by
\begin{equation}
    \hat{R}_{\ast} = floor(\frac{1}{k}{\sum^k_{i=1}{R_{\ast, i}}}),
\end{equation}
where $R_{\ast, i}$ is a rank element of the $i$-th individual and $floor$ denotes the rounding down operation. The interest region is around the $\hat{R}_{\ast}$.

\subsubsection{Progressive Phase}

Progressive phase is used to determine the next search space as shown in Fig. ~\ref{fig:progressive_evolve}. 
At the begining of the PSTRN,
we first obtain initial search space via sampling from the state space at equal intervals as below:
\begin{equation}
    \{R_{min}+b_{1}, R_{min}+2b_{1}, \dots, R_{min}+nb_{1}\}, 
\end{equation}
where $R_{min}$ is the minimum of rank element candidates, and $b_{1}$ is the initial sampling interval.
Then through carrying out evolutionary phase within initial search space, we derive the promising rank
\begin{equation}
    {\hat{\rm\mathbf{R}}} = \{\hat{R}_{0,1}, \hat{R}_{1,1}, \dots, \hat{R}_{d-1,1}\},
\end{equation}
where $\hat{R}_{i,j}, i \in \{0, 1, 2, \dots, d-1\}, j \in \{2, 3, \dots, P\}$ denotes the $i$-th promising rank element in $j$-th evolutionary phase.
Based on the optimized rank, our PSTRN shrinks bound of search space to:
\begin{itemize}
\centering
    \item[-] Low Bound: $\min(\hat{R}_{i,j-1}-s_{j}, R_{min})$,
    \item[-] High Bound: $\max(\hat{R}_{i,j-1}+s_{j}, R_{max})$,
\end{itemize}
where $R_{max}$ is the maximum of rank element candidates, and $\{s_j | j\in \{2, 3, \dots , P\}\}$ is the offset coefficient and usually sets to $b_{j-1}$.
Thus the rank element candidates of the next search space can be expressed as:
\begin{equation}
     \{\hat{R}_{i,j-1}-s_{j}+b_{j}, \hat{R}_{i,j-1}-s_{j}+2b_{j}, \dots,\\
    \hat{R}_{i,j-1}-s_{j}+nb_{j}\},
\label{eq:space}
\end{equation}
where $b_{j}$ is the sampling interval of the $j$-th progressive phase, satisfying
\begin{equation}
    b_{j+1} \leq b_j, j \in \{1, 2, \dots, P\}.
\end{equation}

The interval $b_{j}$ is gradually reduced, and
when
$b_{j}$ decreases to $1$, the progressive phase will stop.

In addition, considering the above Hypothesis~\ref{hyp} cannot be proved by theory, the progressive genetic algorithm may fall into local optima. Therefore, we adds an exploration mechanism to the algorithm. Concretely speaking, except for the initial phase, the algorithm has a $10\%$ probability to choose rank within the search space in the previous evolutionary phase.

In the above evolutionary phase, the solution domain is one of the key components. Generally speaking, it will try to cover all possible states. 
Such an excessive solution domain may lead to the divergence of search algorithm. 
Compared with full state space, our algorithm can improve the search process in computational complexity significantly.

\subsubsection{Weight Inheritance}
During evolutionary phase, to validate the performance, the searched TRN needs to be fully trained, which is the most time-consuming part of the search process.
On MNIST, we can train searched TR-LeNet5 from scratch because of its fast convergence. But the training speed is slow on ResNets.
Thus we employ weight inheritance as a performance estimation acceleration strategy, which is inspired by the architecture evolution~\cite{DBLP:conf/icml/RealMSSSTLK17}.

In our algorithm, to inherit trained weight directly, the rank $R=\{R^k_i | i \in [0, 1, \dots, d-1]\}$ of $k$-th layer needs to follow
\begin{equation}
    R^k_0=R^k_1=\dots=R^k_{d-1}=V_k.
\end{equation}

Obviously,
the number of rank elements to be searched is reduced to $k$ from $d$.
For the $k$-th layer, we will load the checkpoint whenever possible.
Namely, if the $k$-th layer matches $V_k$, the weights are preserved.
Such a method is called warm-up.

During search process, we directly inherit the weights trained in warm-up stage and fine-tune the weights for each searched TRN. 
Instead of training from scratch, fine-tuning the trained weights can greatly resolve the time-consuming problem. 
For example, training ResNet20 on CIFAR10 from scratch requires about 200 epochs. On the contrary, our training with fine-tuning only needs 1 epoch, which brings the acceleration of 200$\times$.

\section{Experiments}
In this section, we conduct experiments to verify the effectiveness of PSTRN. First, to display the relation between the rank elements and performance of TR-based models, we design the synthetic experiment. Then we estimate the effect of the searched TR-based models on prevalent benchmarks.
The optimization objectives of 
NSGA-II~\cite{DBLP:journals/tec/DebAPM02} are classification performance and compression ratio, namely PSTRN-M. In addition, to gain the TR-based model with high performance, we also conduct optimization algorithm PSTRN-S that only consider classification accuracy. All the experiments are implemented on Nvidia Tesla V100 GPUs.
\footnote{TR-based Models can be found at https://github.com/tnbar/tednet. In the initial experiments (TR-LeNet5 on MNIST/Fashion MNIST,  TR-ResNets on CIFAR10/100 and TR-LSTM on HMDB51), the accuracy in search process was obtained on the test dataset instead of the validation dataset which is widely used in works of NAS~\cite{DBLP:conf/iclr/ZophL17}. Thus we supplemented the relevant experiments and shown them in the Appendix.}

\begin{figure}[t]
\centering
    \includegraphics[width=1.05\columnwidth]{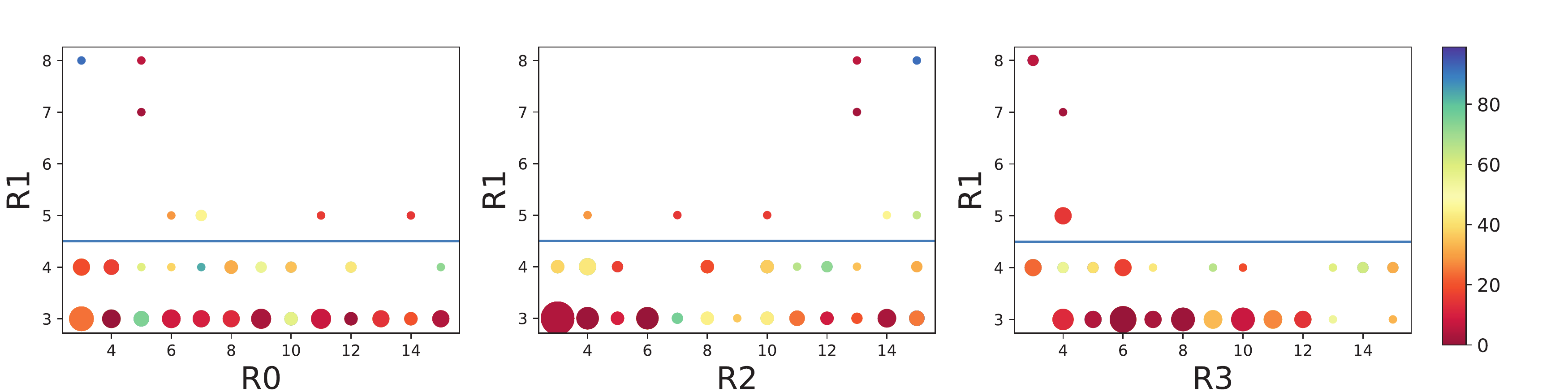}
\caption{Rank distribution between $R_1$ and other rank elements $R_0$, $R_2$, $R_3$ of top-100 models. The size of the circle denotes the number of models that has the same two rank elements, and the circle color represents ranking. The blue line is the border of the interest region.}
\label{fig:distribution}
\end{figure}

\subsection{Synthetic Experiment}
Previous works of rank search lack of heuristic method, so they derive rank elements depending on decomposition, which limits the exploration of searching rank. Hypothesis~\ref{hyp} would bring a promising way to solve this problem, and we would like show the phenomenon of interest region in a synthetic experiment. 

{\bf Experimental Setting} Given a low-rank weight matrix $\textbf{W} \in \mathbb{R}^{144\times 144}$. We generate 5000 samples, and each dimension follows the normal distribution, i.e. $x \sim \mathcal{N}(0, 0.05\textbf{I})$, where $\textbf{I} \in \mathbb{R}^{144}$ is the identity matrix. Then we generate the label $y$ according to $y = \textbf{W}(x+\epsilon)$ for each $x$, where $\epsilon \sim \mathcal{N}(0, 0.05\textbf{I})$ is a random Gaussian noise. 
Data pairs of $\{x, y\}$ constitute the dataset. We divide it into 4000 samples as the training set and 1000 samples as the testing set. 
For the model, we constructed the TR-linear model by replacing the $\textbf{W} \in \mathbb{R}^{144 \times 144}$ with a TRF $\in \mathbb{R}^{12 \times 12 \times 12 \times 12}$. Then we train the TR-linear model with different ranks to completion, and validate the performance on the testing set with Mean-Square Error (MSE) between the prediction $\hat{y}$ and label $y$.  
The rank is denoted as ${\rm\mathbf{R}} = \{R_0, R_1, R_2, R_3 | R_{\ast} \in \{3, 4, \dots, 15\}\}$. 

In the experiment, optimizer sets to Adam with a learning rate $1e-2$, MSE is adopted as loss function and batch size is 128. The total epoch is 100, and 
the learning rate decreases 90\% every 30 epoch.
For a comparison, we run the enumerating results as the baseline, which needs $13^4=28561$ times training. 

\begin{figure}[t]
\centering
\subfloat[]{
    \includegraphics[width=0.45\columnwidth]{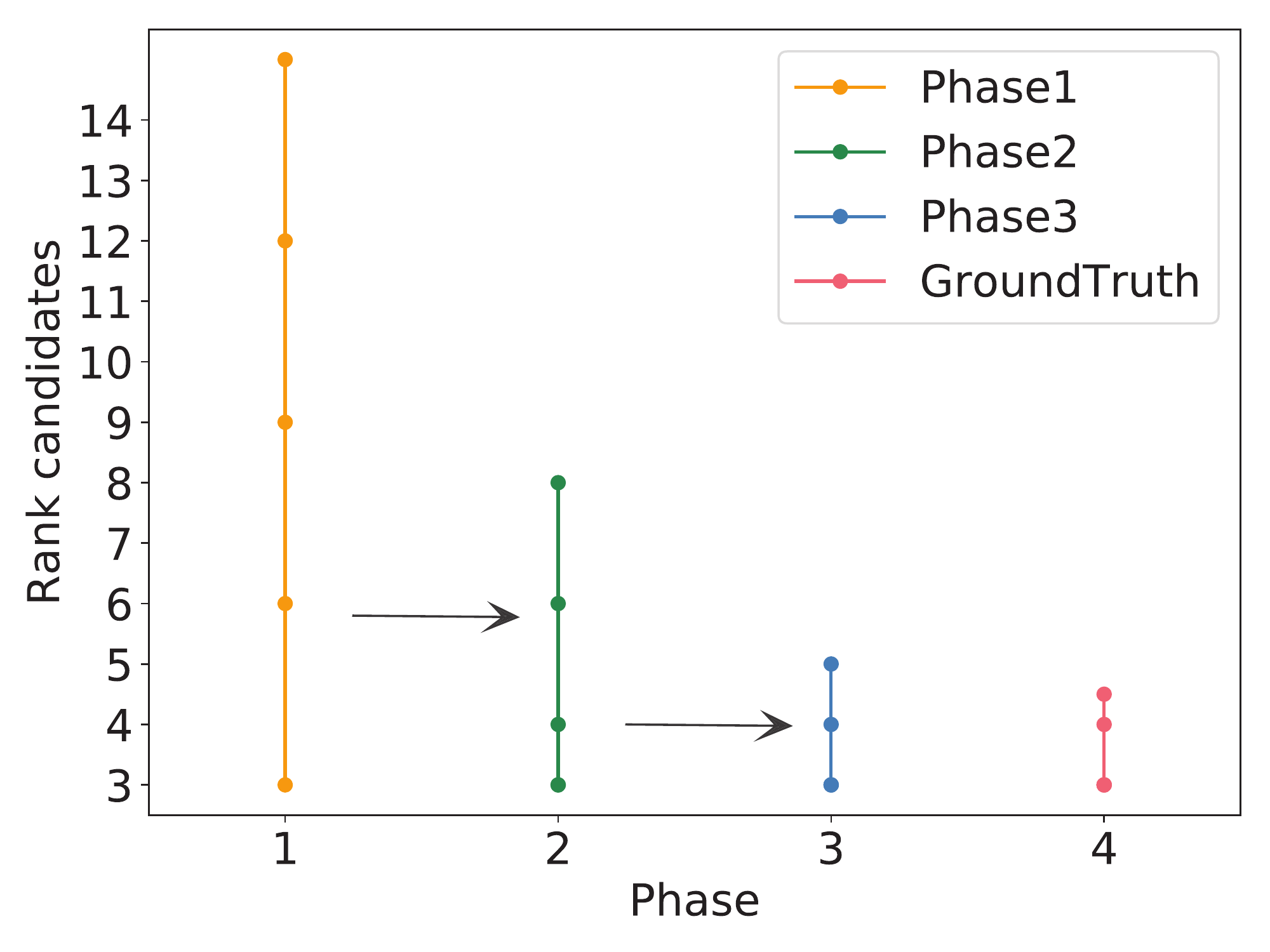}
    \label{fig:converge}
}
\subfloat[]{
    \includegraphics[width=0.45\columnwidth]{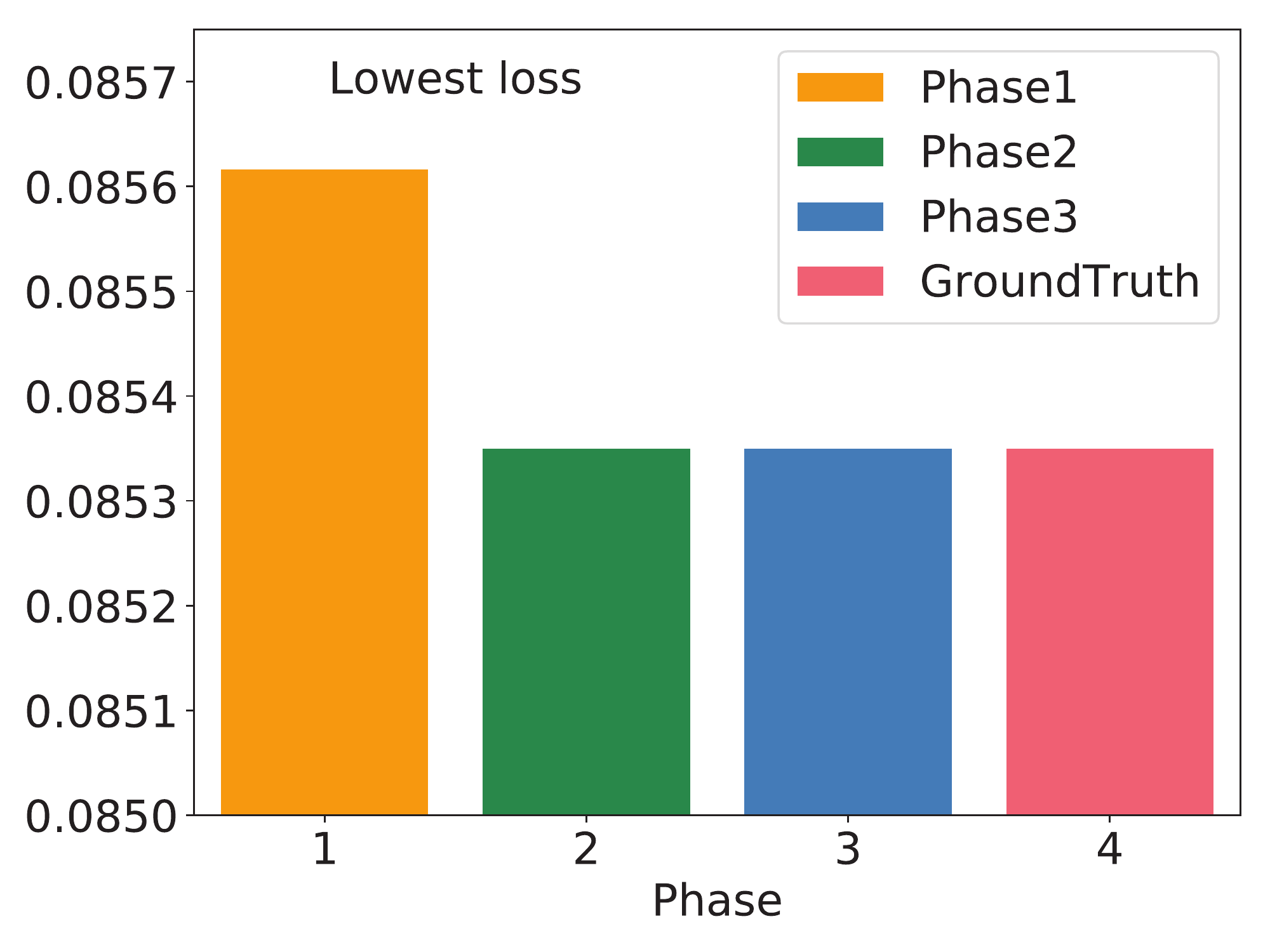}
    \label{fig:performance}
}
\caption{\protect\subref{fig:converge} Interest region approximation of three phases, and groundtruth is the interest region. \protect\subref{fig:performance} Rank performance of models. There are lowest losses of first phase, second phase, third phase and groundtruth from left to right sequentially.}
\label{fig:mlp}
\end{figure}

{\bf Experimental Results} Fig. \ref{fig:distribution} shows the distribution of top-100 rank elements sorted by the value of loss. The size of the circle denotes the number of models who has the same two rank elements. 
And the circle color represents ranking.
It shows that it is not ideal to set each rank element the same arbitrarily.
We calculate the mean $\mu$(3.6) and standard variance $\delta$(0.96) of top-100 models, and derive the interest region $[\mu - \delta, \mu + \delta]$($[2.64, 4.56]$).
Obviously, $R_1$ mostly distributes in the interest region. Should be noted that other rank elements do not show an apparent phenomenon, for the reason that they do not play a critical role in the performance.
Our model can find the interest region that is important to the ability of models and achieve good results.

As shown in Fig.~\ref{fig:converge}, the approximation of interest region gradually approaches groudtruth, which demonstrates that PSTRN can locate interest region precisely.
As illustrated in Fig.~\ref{fig:performance}, our model can find the best rank in the second phase, which proves the powerful capacity of PSTRN algorithm. 
Compared with 28561 enumerating results, we only need $n\_gen \times pop\_size \times P=10 \times 20 \times 3=600$ times training, which is much smaller. $pop\_size$ and $n\_gen$ are the population size and the number of generations respectively.
Undoubtedly, our PSTRN can find the optimal rank efficiently and precisely.

To prove the performance of the progressive search, we also conduct an ablation experiment to compare PSTRN and NSGA-II. $pop\_size$ of NSGA-II is set to 20, which is the same as PSTRN. And $n\_gen$ is set to 30. The experimental results are shown in Table \ref{ablation}. The ranking of the searched model among all 28561 possible models is shown in the last column. It can be seen that our proposed progressive evolutionary algorithm can converge faster.

\begin{table}

\caption{The comparison of PSTRN and NSGA-II.}
\label{ablation}
\centering
\begin{tabular}{cccc}
\hline
\textbf{Phase}     &\textbf{Model}   & \textbf{Generation} & \textbf{Ranking}  \\
\hline
\multirow{2}{*}{1} &PSTRN   &10    & 31      \\
  &NSGA-II & 10     & 32      \\
\multirow{2}{*}{2} &PSTRN  & 20     &  1       \\
  &NSGA-II & 20    &  26       \\
\multirow{2}{*}{3} &PSTRN   & 30    &  1       \\
  &NSGA-II & 30     & 26     \\
\hline
\end{tabular}
\end{table}

\subsection{Experiments on MNIST and Fashion MNIST}
MNIST dataset has 60k grayscale images with 10 object classes, and the resolution of each data is $28\times28$.
Fashion MNIST is more complicated and easy to replace MNIST in experiments.

{\bf Experimental Setting} We evaluate our PSTRN on MNIST and Fashion MNIST by searching TR-LeNet5 that is proposed by Wenqi et al.~\cite{DBLP:conf/cvpr/WangSEWA18}. As shown in Table \ref{TR-LeNet-5}, TR-LeNet5 is constructed with two tensor ring convolutional layers and two tensor ring fully-connected layers. Thus, the total rank is $R = \{R_0, R_1, \dots, R_{19} | R_{\ast} \in \{2, 3, \dots, 30\}\}$. Accordingly, the computational complexity size is $29^{20}\approx1.77\times10^{29}$ for enumeration.

TR-LeNet5 is trained by Adam~\cite{DBLP:journals/corr/KingmaB14} on mini-batches of size 128. The random seed is set to 233. The loss function is cross entropy. Models are trained for 20 epochs with an initial learning rate as 0.002 that is decreased by 0.9 every 5 epochs. 
PSTRN runs 40 generations at each evolutionary phase with population size 30. The number of rank elements searched is 20.  The number of progressive phase is 3. The interval $b_{\ast}$ of each phase is 5, 2 and 1 respectively. Thus complexity of our PSTRN is $n\_gen\times pop\_size \times P=30 \times 40 \times 3=3600$. 

\begin{table}[ht]
\renewcommand\arraystretch{1.2}
\centering
\caption{Dimension of LeNet5 and TR-LeNet5.}
\label{TR-LeNet-5}
\begin{tabular*}{12cm}{lccc}
\hline
   & \textbf{LeNet5} & \multicolumn{2}{c}{\textbf{TR-LeNet5}}  \\

\hline
    layer  & shape & shape & searching rank \\
\hline
conv1  & $5\times5\times1\times20$ & $5\times5\times1\times(4\times5)$ & $R_{0}, R_{1}, R_{2}, R_{3}$ \\
conv2  & $5\times5\times20\times50$ & $5\times5\times(4\times5)\times(5\times10)$ & $R_{4}, R_{5}, R_{6}, R_{7}, R_{8}$ \\
fc1  & $1250\times320$ & $(5\times5\times5\times10)\times(5\times8\times8)$ & $R_{9},R_{10},R_{11},R_{12},R_{13},R_{15},R_{15}$ \\
fc2  & $320\times10$ & $(5\times8\times8)\times10$ & $R_{16}, R_{17}, R_{18}, R_{19}$ \\
\hline
\end{tabular*}
\end{table}

\begin{table}[ht]
\renewcommand\arraystretch{1.2}
\centering
\caption{Comparison with state-of-the-art results for LeNet5 on MNIST.}
\label{MNIST}
\begin{tabular*}{8.3cm}{lccc}
\hline
\textbf{Model} & \textbf{Error(\%)} & \textbf{Params(K)} & \textbf{CR}\\
\hline
Original               & 0.79          & 429         & 1              \\
TR-Nets($r=10$)            & 1.39          & 11          & 39   $\times$  \\
TR-Nets($r=20$)            & 0.69          & 41          & 11   $\times$  \\
TR-Nets($r=30$)$^{ri}$     & 0.70          & 145         & 3    $\times$  \\
\hline
BAMC                   & 0.83          & -           & -      \\
LR-L                   & 0.75          & 27          & 15.9 $\times$  \\
\hline
\textbf{PSTRN-M(Ours)} & 0.57          & \textbf{26} & 16.5 $\times$  \\
\textbf{PSTRN-S(Ours)} & \textbf{0.49} & 66          & 6.5  $\times$  \\
\hline
\end{tabular*}
\end{table}

\begin{figure}[t]
\centering
\includegraphics[width=0.9\columnwidth]{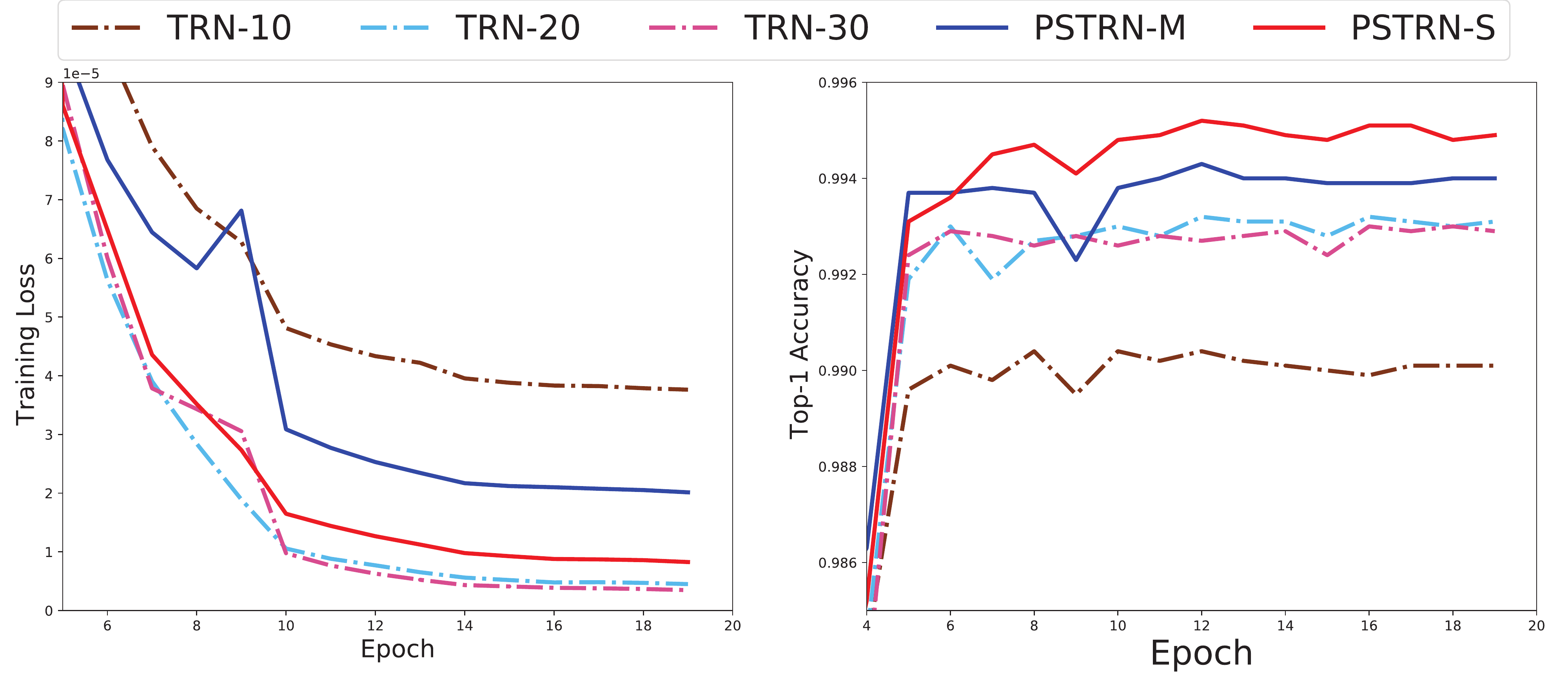}
\caption{Training process on MNIST. TRN-10/20/30 are TR-Nets that set rank to 10, 20 and 30 respectively. TRN-20 and TRN-30 are over-fitting.}
\label{fig:mnist}
\end{figure}

{\bf Experimental Results} Experimental results are summarized in Table \ref{MNIST} and Table~\ref{Fashion MNIST}, where original LeNet5 is proposed by LeCun et al.~\cite{lecun1998gradient}, Bayesian Automatic Model Compression(BAMC)~\cite{wang2020bayesian} leverages Dirichlet process
mixture models to explore the layer-wise quantization policy, LR-L~\cite{DBLP:conf/cvpr/IdelbayevC20} learn the rank of each layer for SVD, and TR-Nets~\cite{DBLP:conf/cvpr/WangSEWA18} compresses deep neural network via tensor ring decomposition with equal rank elements. The superscript ${ri}$ represents there are results of re-implement, and $r$ is the rank of works that set rank elements to equal.
These settings would be retained in subsequent experiments. Both of the search processes for MNIST and Fashion MNIST cost about 5 GPU days.
In Table \ref{MNIST}, the first block shows the results of rank-fixed methods, which manually set rank elements to equal. The second block is the work that automatically compresses the model. As expected, both PSTRN-M and PSTRN-S achieve best performance on MNIST. Our algorithm compress LeNet5 with compression ratio as 6.5x and 0.49\% error rate. And our PSTRN can also exceed models that set rank manually on Fashion MNIST as shown in Table~\ref{Fashion MNIST}.
Further, Fig. \ref{fig:mnist} demonstrates that our proposed approach outperforms rank-fixed works on MNIST.
Obviously, when fixed rank $r$ is bigger than 20, the TR-Nets will be over-fitting. And our proposed work can find the suitable rank with best performance. 

The ranks of searched TR-LeNet5 are shown in Table~\ref{TR-LeNet-5-rank}. The symbol $//$ indicates the different layers.

\begin{table}[ht]
\renewcommand\arraystretch{1.2}
\centering
\caption{Results on Fashion MNIST.}
\label{Fashion MNIST}
\begin{tabular*}{8.4cm}{lccc}
\hline
\textbf{Model} & \textbf{Error(\%)} & \textbf{Params(K)} & \textbf{CR}\\
\hline
Original$^{ri}$         & 7.4          & 429    & 1             \\
TR-Nets($r=10$)$^{ri}$      & 9.63         & 16     & 26.5 $\times$ \\
TR-Nets($r=20$)$^{ri}$      & 8.67         & 65     & 6.6  $\times$ \\
TR-Nets($r=30$)$^{ri}$      & 8.64         & 145    & 3.0  $\times$ \\
\hline
\textbf{PSTRN-M(Ours)}  & 8.05          & 49     & 8.8  $\times$ \\
\textbf{PSTRN-S(Ours)}  & \textbf{7.85} & 62     & 6.9  $\times$ \\
\hline
\end{tabular*}
\end{table}

\begin{table}[ht]
\renewcommand\arraystretch{1.2}
\centering
\caption{Rank of searched TR-LeNet5.}
\label{TR-LeNet-5-rank}
\begin{tabular*}{12cm}{lccc}
\hline
\textbf{Model} & \textbf{Error(\%)} & \textbf{Rank} \\
\hline
\multicolumn{3}{c}{MNIST} \\
\hline
PSTRN-M & 0.57 & \{6,20,14,8//12,20,2,20,16//16,20,12,12,8,6,26//8,2,6,20\}   \\
PSTRN-S & 0.49 & \{2,24,18,8//8,30,18,30,22//26,22,26,30,14,30,8//24,12,10,30\}   \\
\hline
\multicolumn{3}{c}{Fashion MNIST} \\
\hline
PSTRN-M & 8.05 & \{8,14,8,18//14,18,22,28,6//24,20,14,20,20,10,22//20,20,6,20\}   \\
PSTRN-S & 7.85 & \{22,12,6,22//16,22,30,22,16//18,24,30,30,8,30,24//22,20,6,18\}  \\
\hline
\end{tabular*}
\end{table}

\begin{table}[ht]
\renewcommand\arraystretch{1.2}
\centering
\caption{Dimension of ResNet and TR-ResNet. $\Psi$ denotes the number of blocks in units. $\Psi$ is set to 3 and 5 to represent ResNet-20 and ResNet-32 respectively.}
\label{TR-ResNet}
\begin{tabular*}{12cm}{lccc}
\hline
   & \textbf{ResNet} & \multicolumn{2}{c}{\textbf{TR-ResNet} } \\
\hline
layer  & shape & shape & searching rank \\
\hline
conv1  & $3\times3\times3\times16$ & $9\times3\times(4\times2\times2)$ & $R_{0}$ \\
unit1  &$ResBlock(3,16,16)\times \Psi$ & $9\times(4\times2\times2)\times(4\times2\times2)$& $R_{1}$ \\
unit2  & $ResBlock(3,16,32)$ &  $9\times(4\times2\times2)\times(4\times4\times2)$ & $R_{2}$ \\
 & $ResBlock(3,32,32) \times (\Psi-1)$ &  $9\times(4\times4\times2)\times(4\times4\times2)$ & $R_{3}$ \\
unit3  & $ResBlock(3,32,64)$ & $9\times(4\times4\times2)\times(4\times4\times4)$ & $R_{4}$ \\
   & $ResBlock(3,64,64)\times (\Psi-1)$ & $9\times(4\times4\times4)\times(4\times4\times4)$  &$R_{5}$ \\
fc1  & $64\times10$ & $(4\times4\times4)\times10$ & $R_{6}$ \\

\hline
\end{tabular*}
\end{table}

\begin{table}[ht]
\renewcommand\arraystretch{1.2}
\centering
\caption{Dimension of WideResNet28-10 and TR-WideResNet28-10.}
\label{TR-WideResNet}
\begin{tabular*}{12.5cm}{lccc}
\hline
   & \textbf{WideResNet28-10} & \multicolumn{2}{c}{\textbf{TR-WideResNet28-10}}  \\
\hline
layer  & shape & shape & searching rank \\
\hline
conv1 & $3\times3\times3\times16$ & 9$\times3\times(4\times2\times2)$ & $R_{0}$ \\
unit1  & $ResBlock(3,16,160)$ & $9\times(4\times2\times2)\times(10\times4\times2\times2)$& $R_{1}$ \\
   & $ResBlock(3,160,160)\times 3$ & $9\times(10\times4\times2\times2)\times(10\times4\times2\times2)$& $R_{2}$ \\
unit2 & $ResBlock(3,160,320)$&  $9\times(10\times4\times2\times2)\times(10\times4\times4\times2)$ & $R_{3}$ \\
 & $ResBlock(3,320,320) \times 3$ &  $9\times(10\times4\times4\times2)\times(10\times4\times4\times2)$ & $R_{4}$ \\
unit3  & $ResBlock(3,320,640)$ & $9\times(10\times4\times4\times2)\times(10\times4\times4\times4)$ & $R_{5}$\\
   & $ResBlock(3,640,640)\times 3$ & $9\times(10\times4\times4\times4)\times(10\times4\times4\times4)$  & $R_{6}$\\
fc1 & $640\times10$ & $(10\times4\times4\times4)\times10$ & $R_{7}$ \\

\hline
\end{tabular*}
\end{table}

\begin{table}[ht]
\renewcommand\arraystretch{1.2}
\centering
\caption{Comparison with state-of-the-art results for ResNet20 on CIFAR. }
\label{ResNet20}
\begin{tabular*}{8.4cm}{lccc}
\hline
\textbf{Model} & \textbf{Error(\%)} & \textbf{Params(M)} & \textbf{CR}\\
\hline
\multicolumn{4}{c}{CIFAR10} \\
\hline
Original               & 8.75           & 0.27   & 1  \\
TR-Nets($r=10$)            & 12.5           & 0.05   & 5.40 $\times$ \\
TR-RL                 & 11.7           & 0.04   & 6.75 $\times$ \\
LR-L                   & 12.89          & 0.05   & 5.40 $\times$ \\
\textbf{PSTRN-M(Ours)} & \textbf{10.7} & 0.04   & 6.75 $\times$ \\
\hline
LR-L                   & 9.49           & 0.11   & 2.45 $\times$ \\
TR-Nets($r=15$)$^{ri}$     & 9.22           & 0.13   & 2.08

$\times$  \\
\textbf{PSTRN-S(Ours)} & \textbf{9.20}  & 0.12   & 2.25 $\times$ \\
\hline
\multicolumn{4}{c}{CIFAR100} \\
\hline
Original               & 34.60          & 0.28   & 1  \\
TR-Nets($r=10$)$^{ri}$     & 36.45      & 0.07   & 4 $\times$ \\
\textbf{PSTRN-M(Ours)} & \textbf{36.38}          & 0.07   & 4 $\times$ \\
\hline
TR-Nets($r=15$)$^{ri}$     & 34.49          & 0.15   & 1.9 $\times$ \\
\textbf{PSTRN-S(Ours)} & \textbf{33.87}          & 0.13   & 2.2 $\times$ \\
\hline
\end{tabular*}
\end{table}

\begin{table}[ht]
\renewcommand\arraystretch{1.2}
\centering
\caption{Comparison with state-of-the-art results for ResNet32 on CIFAR.}
\label{ResNet32}
\begin{tabular*}{8.4cm}{lccc}
\hline
\textbf{Model} & \textbf{Error(\%)} & \textbf{Params(M)} & \textbf{CR}\\
\hline
\multicolumn{4}{c}{CIFAR10} \\
\hline
Original               & 7.5           & 0.46     & 1      \\
Tucker                 & 12.3          & 0.09     & 5.1 $\times$ \\
TT($r=13$)             & 11.7          & 0.10     & 4.8 $\times$  \\
TR-Nets($r=10$)            & 9.4           & 0.09     & 5.1 $\times$   \\
TR-RL                 & 11.9          & 0.03     & 1.5 $\times$   \\
LR-L                   & 10.56         & 0.09     & 5.1 $\times$   \\
\textbf{PSTRN-M(Ours)} & \textbf{9.4}  & 0.09     & 5.1 $\times$   \\
\hline
TR-Nets($r=15$)$^{ri}$     & 8.76          & 0.21     & 2.2 $\times$   \\
\textbf{PSTRN-S(Ours)} & \textbf{8.56} & 0.18     & 2.6 $\times$   \\
\hline
\multicolumn{4}{c}{CIFAR100} \\
\hline
Original               & 31.90          & 0.47     & 1  \\
Tucker                 & 42.20        & 0.09     & 5.1 $\times$ \\
TT($r=13$)             & 37.10          & 0.10     & 4.6 $\times$  \\
TR-Nets($r=10$)            & 33.30          & 0.097    & 4.8 $\times$ \\
\textbf{PSTRN-M(Ours)} & \textbf{33.23} & 0.094    & 5.2 $\times$ \\
\hline
TR-Nets($r=15$)$^{ri}$       & 32.73         & 0.227    & 2.1 $\times$ \\
\textbf{PSTRN-S(Ours)} & \textbf{31.95} & 0.210    & 2.2 $\times$ \\
\hline
\end{tabular*}
\end{table}

\begin{table}[ht]
\renewcommand\arraystretch{1.2}
\centering
\caption{Comparison with state-of-the-art results for WideResNet28-10 on CIFAR.}
\label{WideResNet28}
\begin{tabular*}{8.4cm}{lccc}
\hline
\textbf{Model} & \textbf{Error(\%)} & \textbf{Params(M)} & \textbf{CR}\\
\hline
Original               &    5.0        &   36.2   &  1   \\
Tucker                 & 7.8      & 6.7     & 5$\times$\\
TT($r=13$)            & 8.4        & 0.18 & 154$\times$  \\
TR-Nets($r=10$)           & 7.3          &   0.15    & 173$\times$   \\
TR-Nets($r=15$)    &   7.0      &  0.30  &  122$\times$  \\
\textbf{PTRNS-M(Ours)} &  \textbf{6.9}  &   0.26   &   141$\times$  \\

\hline
\end{tabular*}
\end{table}

\subsection{Experiments on CIFAR10 and CIFAR100}
Both CIFAR10 and CIFAR100 datasets consist of 50,000 train images and 10,000 test images with size as $32 \times 32 \times 3$. CIFAR10 has 10 object classes and CIFAR100 has 100 categories.

{\bf Experimental Setting} The dimension of TR-RseNet is shown in Table\ref{TR-ResNet}, $\Psi$ is the number of ResBlock. TR-ResNet32 is built as introduced by Wenqi et al.~\cite{DBLP:conf/cvpr/WangSEWA18} with $\Psi$ as 5, and TR-ResNet20 is constructed as proposed by Zhiyu et al.~\cite{DBLP:conf/icassp/ChengLFB20} with $\Psi$ as 3.
First, we apply the PSTRN-M/S to search TR-ResNet20/32 on CIFAR10.
Further, we transfer TR-ResNet20/32 searched by PSTRN-M/S on CIFAR10 into CIFAR100 to evaluate the transferability of PSTRN.  
Considering that training TR-ResNet20/32 on CIFAR10 is time-consuming, we apply the weight inheritance to accelerate the process. 
Specifically, we pre-train weight of the model in warm-up stage. Then we load pre-trained weights directly.
The training epoch of warm-up is set to 30. The rank $R = \{R_0, R_1, \dots, R_{6} | R_{\ast} \in \{2, 3, \dots, 20\}\}$.

TR-ResNets are trained via SGD~\cite{ruder2016overview} with momentum 0.9 and a weight decay of $5\times 10^{-4}$ on mini-batches of size 128. The random seed is set to 233. The loss function is cross entropy. TR-ResNets are trained for 200 epochs with an initial learning rate as 0.02 that is decreased by 0.8 after every 60 epochs. 
Our approach runs 20 generations at each evolutionary phase with population size 30. The number of rank elements searched is 7. The number of progressive phase is 3. The interval $b_{\ast}$ of each phase is 3, 2 and 1 respectively. The complexity of our approach is $n\_gen \times pop\_size \times P=30 \times 40 \times 3=3600$, which is much smaller than computational complexity $19^{7}\approx8.9 \times 10^{8}$ of enumeration method.

To better validate our algorithm, we also apply the proposed algorithm to TR-WideResNet28-10 on CIFAR10. The experimental setting is exactly same as TR-ResNet. The dimension of TR-WideResNet28-10 is shown as Table \ref{TR-WideResNet}. The rank $R = \{R_0, R_1, \dots, R_{7} | R_{\ast} \in \{2, 3, \dots, 20\}\}$.

{\bf Experimental Results} The results for ResNet20 and ResNet32 are given in Table \ref{ResNet20} and Table \ref{ResNet32}. The search process for ResNet20/32 and WideResNet28-10 cost about 2.5/3.2 GPU days and 3.8 GPU days respectively.
In the Tables \ref{ResNet20} and \ref{ResNet32}, original ResNet20/32 are the model proposed by Kaiming et al.~\cite{DBLP:conf/cvpr/HeZRS16}. Tucker~\cite{DBLP:journals/corr/KimPYCYS15} and TT~\cite{garipov2016ultimate} are works that compress neural networks by other tensor decomposition methods. 
TR-RL~\cite{DBLP:conf/icassp/ChengLFB20} search the rank of TR-based model based on reinforcement learning.
The first block compares PSTRN-M with the results of low rank decomposition works that have fewer parameters. Obviously, PSTRN-M 
surpasses other methods in both classification accuracy and compression ratio. The second block reports the performance of PSTRN-S and models that are poor on compression. Results tell that our
algorithm achieve high performance and beyonds works with 0.10+M parameters. 

The results for TR-WideResNet28-10 are given in Table \ref{WideResNet28}. Apparently, PSTRN-M exceeds other methods in both classification accuracy and compression ratio. The experimental results show that PSTRN-M can exceed the manually designed models in terms of classification accuracy and compression ratio. In the case that the number of parameters is smaller than the manually designed model with $r$ as 15, the accuracy of PSTRN-M is higher.
\begin{table}[ht]
\renewcommand\arraystretch{1.2}
\centering
\caption{Rank of searched TR-ResNet20.}
\label{TR-ResNet20-rank}
\begin{tabular*}{9.5cm}{lccc}
\hline
\textbf{Model} & \textbf{Error(\%)} & \textbf{Rank} \\
\hline
\multicolumn{3}{c}{CIFAR10} \\
\hline
PSTRN-M & 10.7 & \{ 4 // 8 // 6 // 8 // 6 // 12 // 10 \}    \\
PSTRN-S & 9.2  & \{ 8 // 15 // 10 // 12 // 14 // 18 // 12 \}     \\
\hline
\multicolumn{3}{c}{CIFAR100} \\
\hline
PSTRN-M & 36.38 & \{ 8 // 10 // 6 // 8 // 8 // 12 // 12 \}    \\
PSTRN-S & 33.87  & \{ 8 // 15 // 10 // 12 // 14 // 18 // 12 \}     \\
\hline
\end{tabular*}
\end{table}

\begin{table}[ht]
\renewcommand\arraystretch{1.2}
\centering
\caption{Rank of searched TR-ResNet32.}
\label{TR-ResNet32-rank}
\begin{tabular*}{9.5cm}{lccc}
\hline
\textbf{Model} & \textbf{Error(\%)} & \textbf{Rank} \\
\hline
\multicolumn{3}{c}{CIFAR10} \\
\hline
PSTRN-M & 9.4 & \{ 3 // 12 // 9 // 9 // 9 // 9 // 6 \}    \\
PSTRN-S & 8.56  & \{ 17 // 14 // 10 // 13 // 18 // 15 // 10 \}     \\
\hline
\multicolumn{3}{c}{CIFAR100} \\
\hline
PSTRN-M & 33.23 & \{ 4 // 8 // 6 // 8 // 12 // 12 // 10 \}    \\
PSTRN-S & 31.95  & \{ 16 // 14 // 12 // 14 // 20 // 16 // 8 \}     \\
\hline
\end{tabular*}
\end{table}

\begin{table}[ht]
\renewcommand\arraystretch{1.2}
\centering
\caption{Rank of searched TR-WideResNet28-10.}
\label{TR-WideResNet28-rank}
\begin{tabular*}{9.8cm}{lccc}
\hline
\textbf{Model} & \textbf{Error(\%)} & \textbf{Rank} \\
\hline
PSTrRN-M & 6.9 & \{ 8 // 12 // 8 // 11 // 12 // 14 // 18 // 9 \}    \\
\hline
\end{tabular*}
\end{table}

In addition, through transferring the searched PSTRN-M/S on CIFAR10 into CIFAR100, PSTRN obtain excellent results as well. This proves that our proposed PSTRN can not only find well-performed models, but also possesses transferability. 

The ranks of searched TR-ResNet20 and TR-ResNet32 are shown in Table~\ref{TR-ResNet20-rank}, ~\ref{TR-ResNet32-rank}, and Table~\ref{TR-WideResNet28-rank}. 

\subsection{Experiments on HMDB51 and UCF11}
HMDB51 dataset is a large collection of realistic videos from various sources, such as movies and web videos. The dataset is composed of 6766 video clips from 51 action categories. UCF11 dataset contains 1600 video clips of a resolution $320 \times 240$ and is divided into 11 action categories. Each category consists of 25 groups of videos, within more than 4 clips in one group. 

{\bf Experimental Setting} In this experiment, we sample 12 frames from each video clip randomly.
And then we extract features from the frames via Inception-V3~\cite{DBLP:conf/cvpr/SzegedyVISW16} input vectors and reshape it into $64 \times 32$.
The shape of
hidden layer tensor is set as $32 \times 64 = 2048$. For TR-LSTM, as shown in Table \ref{TR-LSTM}, the rank is denoted as ${\rm\mathbf{R}} = \{R_0, R_1, R_2, R_3 | R_{\ast} \in \{15, 16, \dots, 60\}\}$. The complexity of enumerating approach is $46^{4}\approx4.5\times10^{6}$.

TR-LSTM is trained via Adam with a weight decay of $1.7\times 10^{-4}$ on mini-batches of size 32. The random seed is set to 233. The loss function is the cross-entropy. In the search phase, searched models are trained for 100 epochs with an initial learning rate of 1e-5. Our approach runs 20 generations at each evolutionary phase with a population size of 20. The number of rank elements searched is 4. The number of progressive phases is 3. The interval $b_{\ast}$ of each phase is 8, 3 and 1 respectively. The computational complexity of our PSTRN is $n\_gen \times pop\_size \times P=20 \times 20 \times 3=1200$ that is much smaller.
\begin{table}[ht]
\renewcommand\arraystretch{1.2}
\centering
\caption{Dimension of LSTM and TR-LSTM.}
\label{TR-LSTM}
\begin{tabular*}{9cm}{lccc}
\hline
   &\textbf{LSTM} & \multicolumn{2}{c}{\textbf{TR-LSTM}}  \\

\hline
layer  & shape & shape & searching rank \\
\hline
fc  & $2048\times2048$ & $(64\times32)\times(32\times64)$ & $R_{0}, R_{1}, R_{2}, R_{3}$ \\

\hline
\end{tabular*}
\end{table}

\begin{table}[t]
\renewcommand\arraystretch{1.2}
\centering
\caption{Results of TR decomposition for LSTM on HMDB51.}
\label{HMDB51}
\begin{tabular*}{8.4cm}{lccc}
\hline
\textbf{Model} & \textbf{Error(\%)} & \textbf{Params(M)} & \textbf{CR}\\
\hline
Original               & 51.85          & 17        & 1              \\
TR-LSTM($r=15$)$^{ri}$            & 45.94         & 0.06          & 258.9$\times$  \\
TR-LSTM($r=30$)$^{ri}$            & 41.65         & 0.26          & 64.7$\times$  \\
TR-LSTM($r=50$)$^{ri}$            & 42.25          & 0.73         & 23.3$\times$  \\
TR-LSTM($r=60$)$^{ri}$            & 41.95          & 1.04          & 16.2$\times$  \\
\hline
\textbf{PSTRN-M(Ours)} & 40.33 & 0.36         & 46.7$\times$  \\
\textbf{PSTRN-S(Ours)} & \textbf{39.96} & 0.49          & 34.7$\times$  \\
\hline
\end{tabular*}
\end{table}

\begin{table}[t]
\renewcommand\arraystretch{1.2}
\centering
\caption{Results of TR decomposition for LSTM on UCF11.}
\label{UCF11}
\begin{tabular*}{8.4cm}{lccc}
\hline
\textbf{Model} & \textbf{Error(\%)} & \textbf{Params(M)}& \textbf{CR}\\
\hline
Original              &    12.66       &      17     &         1      \\
TR-LSTM($r=15$)$^{ri}$           &  8.86       &       0.06     &  258.91$\times$ \\
TR-LSTM($r=40$)$^{ri}$            &7.91          & 0.46        & 36.41$\times$ \\
TR-LSTM($r=60$)$^{ri}$            &   7.28       &  1.04       &  16.18$\times$\\
\hline
\textbf{PSTRN-M(Ours)} &7.91  &       0.09     & 190.10$\times$ \\
\textbf{PSTRN-S(Ours)} &\textbf{6.65}  &    0.20       & 84.64$\times$  \\
\hline
\end{tabular*}
\end{table}

\begin{table}[h]
\centering
\caption{Rank of searched TR-LSTM.}
\label{TR-LSTM-rank}
\begin{tabular}{lll}
\hline
\textbf{Model} & \textbf{Error(\%)} & \textbf{Rank} \\
\hline
\multicolumn{3}{c}{HMDB51} \\
\hline
PSTRN-M & 40.33 & \{ 52 // 17 // 34 // 37 \}  \\
PSTRN-S & 39.96 & \{ 45 // 42 // 36 // 42 \}  \\
\hline
\multicolumn{3}{c}{UCF11} \\
\hline
PSTRN-M & 7.91 & \{ 19 // 15 // 20 // 16 \}  \\
PSTRN-S &6.65 &\{ 39 // 19 // 34 // 19 \}  \\
\hline
\end{tabular}
\end{table}

{\bf Experimental Results} The comparisons between our approach and manually-designed method are shown in Table~\ref{HMDB51} and Table~\ref{UCF11}. The search process for HMDB51 and UCF11 cost about 1.4 GPU days and 0.5 GPU days respectively. Experimental results on Table~\ref{HMDB51} show that our searched rank exceed in others with equal rank elements on HMDB51. The results for UCF11 are given in Table~\ref{UCF11}. As can be seen, with the compression ratio greater than twice that of the manually designed model with $r$ as 40, the classification accuracy of PSTRN-S is 1.26\% higher. PSTRN-M also achieves higher accuracy with fewer parameters. Table~\ref{TR-LSTM-rank} demonstrates the ranks for TR-LSTM that searched via PSTRN.

\textbf{Remark} Unlike the Pytorch, Keras is a high level package with many tricks under the table, e.g. hard sigmoid in RNNs. Thus, for fairer comparison and validation of search results, we implement this experiment in Pytorch and remove the tricks in the Keras package. Additionally, through Keras implement, our searched TR-LSTM achieve 64.5\% accuracy with a compression ratio 48, which is better than 63.8\% with a compression ratio 25~\cite{DBLP:conf/aaai/PanXWYWBX19}.

Another important component is the shape of a tensor ring decomposition. The method of choosing the shape is notorious. And actually there are almost not any way to select the shape efficiently. Therefore, our PSTRN simply chooses a shape with a similar size by manipulation. The effect of shape on TR-based model is unknown and waits to be solved in the future.

Generally, the rank has similar attribution in different kinds of tensor decomposition like Tucker, Tensor Train and so on. It is reasonable to assume that the Hypothesis~\ref{hyp} is suitable for them. Therefore, it is promising to employ PSTRN on them to explore their potential power.

\section{Conclusion}
In this paper, we propose a novel algorithm PSTRN based on Hypothesis~\ref{hyp} to search optimal rank. As a result, our algorithm compresses LeNet5 with 16$\times$ compression ratio and 0.22\% accuracy improvement on MNIST. And on CIFAR10 and CIFAR100, our work achieves state-of-the-art performance with a high compression ratio for ResNet20, ResNet32 and WideResNet28-10. Not only the CNN, we also show excellent performance of LSTM on HMDB51 and UCF11. 
Further, for large-scale datasets, we will explore more performance evaluation acceleration strategies to optimize rank elements more efficiently.

\section*{Conflict of Interest Statement}
On behalf of all authors, the corresponding author states that there is no conflict of interest.

\begin{acknowledgements}
This work is supported partly by the National Natural Science Foundation of China (NSFC) under Grants No. 62006226 and the National Key Research and Development Program of China No. 2018AAA0100204.
\end{acknowledgements}

%
%

\bibliographystyle{spmpsci}      


\bibliography{reference}
\appendix

\section{Appendix}

For the following experiments, we randomly split 90\% from the original training dataset for training and 10\% for validation in the search process. And remaining experimental settings are consistent with the text. The experimental results demonstrate that PSTRN can still achieve better performance in terms of accuracy and compression ratio.
  
\subsection{TR-LeNet5 on MNIST and Fashion MNIST}
We search for TR-LeNet5 on MNIST~\cite{DBLP:journals/spm/Deng12} and Fashion MNIST~\cite{DBLP:journals/corr/abs-1708-07747} (10 classes, 60k grayscale images of 28 $\times$ 28). The experimental results are shown in Table~\ref{MNIST_A} and Table~\ref{Fashion MNIST_A}.

\begin{table}[ht]
\renewcommand\arraystretch{1.2}
\centering
\caption{Comparison with state-of-the-art results for LeNet5 on MNIST.}
\label{MNIST_A}
\begin{tabular*}{8.3cm}{lccc}
\hline
\textbf{Model} & \textbf{Error(\%)} & \textbf{Params(K)} & \textbf{CR}\\
\hline
Original               & 0.79          & 429         & 1              \\
TR-Nets($r=10$)            & 1.39          & 11          & 39   $\times$  \\
TR-Nets($r=20$)            & 0.69          & 41          & 11   $\times$  \\
TR-Nets($r=30$)$^{ri}$     & 0.70          & 145         & 3    $\times$  \\
\hline
BAMC                   & 0.83          & -           & -      \\
LR-L                   & 0.75          & 27          & 11.9 $\times$  \\
\hline
\textbf{PSTRN-M(Ours)} & 0.6          & 36 & 16.5 $\times$  \\
\textbf{PSTRN-S(Ours)} & \textbf{0.55} & 68          & 6.3  $\times$  \\
\hline
\end{tabular*}
\end{table}

\begin{table}[ht]
\renewcommand\arraystretch{1.2}
\centering
\caption{Results on Fashion MNIST.}
\label{Fashion MNIST_A}
\begin{tabular*}{8.4cm}{lccc}
\hline
\textbf{Model} & \textbf{Error(\%)} & \textbf{Params(K)} & \textbf{CR}\\
\hline
Original$^{ri}$         & 7.4          & 429    & 1             \\
TR-Nets($r=10$)$^{ri}$      & 9.63         & 16     & 26.5 $\times$ \\
TR-Nets($r=20$)$^{ri}$      & 8.67         & 65     & 6.6  $\times$ \\
TR-Nets($r=30$)$^{ri}$      & 8.64         & 145    & 3.0  $\times$ \\
\hline
\textbf{PSTRN-M(Ours)}  & 8.42          & 50     & 8.6  $\times$ \\
\textbf{PSTRN-S(Ours)}  & \textbf{8.35} & 70     & 6.1  $\times$ \\
\hline
\end{tabular*}
\end{table}

\subsection{TR-ResNets on CIFAR10 and CIFAR100}

We search TR-ResNets on CIFAR10 (10 classes, 60k RGB images of 32 $\times$ 32) and transfer the searched model to CIFAR100~\cite{krizhevsky2009learning} (100 classes, 60k RGB images of 32 $\times$ 32). 
The experimental results are shown in Table~\ref{ResNet20_A} and Table~\ref{ResNet32_A}.
\begin{table}[ht]
\renewcommand\arraystretch{1.2}
\centering
\caption{Comparison with state-of-the-art results for ResNet20 on CIFAR. }
\label{ResNet20_A}
\begin{tabular*}{8.4cm}{lccc}
\hline
\textbf{Model} & \textbf{Error(\%)} & \textbf{Params(M)} & \textbf{CR}\\
\hline
\multicolumn{4}{c}{CIFAR10} \\
\hline
Original               & 8.75           & 0.27   & 1  \\
TR-Nets($r=10$)     & 12.5           & 0.05   & 5.40 $\times$ \\
TR-RL                 & 11.7           & 0.04   & 6.75 $\times$ \\
LR-L                   & 12.89          & 0.05   & 5.40 $\times$ \\
\textbf{PSTRN-M(Ours)} & \textbf{11.1} & 0.05   & 5.40 $\times$ \\
\hline
TR-Nets($r=20$)$^{ri}$     & 9.03           & 0.22   & 1.22 $\times$  \\
\textbf{PSTRN-S(Ours)} & \textbf{8.94}  & 0.16   & 1.69 $\times$ \\
\hline
\multicolumn{4}{c}{CIFAR100} \\
\hline
Original               & 34.60          & 0.28   & 1  \\
TR-Nets($r=10$)$^{ri}$     & 36.45      & 0.07   & 4.7 $\times$ \\
\textbf{PSTRN-M(Ours)} & \textbf{35.9}          & 0.07   & 4.7 $\times$ \\
\hline
TR-Nets($r=20$)$^{ri}$     & 33.01          & 0.26   & 2.0 $\times$ \\
\textbf{PSTRN-S(Ours)} & \textbf{32.93}          & 0.16   & 2.3 $\times$ \\
\hline
\end{tabular*}
\end{table}

\subsection{TR-LSTM on HMDB51}
We search TR-LSTM on HMDB51~\cite{DBLP:conf/iccv/KuehneJGPS11} (51 classes, 6766 video clips). Experimental results are shown in Table~\ref{HMDB51_A}.

\begin{table}[ht]
\renewcommand\arraystretch{1.2}
\centering
\caption{Comparison with state-of-the-art results for ResNet32 on CIFAR.}
\label{ResNet32_A}
\begin{tabular*}{8.4cm}{lccc}
\hline
\textbf{Model} & \textbf{Error(\%)} & \textbf{Params(M)} & \textbf{CR}\\
\hline
\multicolumn{4}{c}{CIFAR10} \\
\hline
Original               & 7.5           & 0.46     & 1      \\
TT($r=13$)             & 11.7          & 0.10     & 4.8 $\times$  \\
TR-Nets($r=10$)            & 9.4           & 0.09     & 5.1 $\times$   \\
LR-L                   & 10.56         & 0.09     & 5.1 $\times$   \\
\textbf{PSTRN-M(Ours)} & \textbf{9.37}  & 0.09     & 5.1 $\times$   \\
\hline
TR-Nets($r=15$)$^{ri}$     & 8.76          & 0.21     & 2.2 $\times$   \\
\textbf{PSTRN-S(Ours)} & \textbf{8.38} & 0.21     & 2.2 $\times$   \\
\hline
\multicolumn{4}{c}{CIFAR100} \\
\hline
Original               & 31.90          & 0.47     & 1  \\
Tucker                 & 42.20        & 0.09     & 5.1 $\times$ \\
TT($r=13$)             & 37.10          & 0.10     & 4.6 $\times$  \\
TR-Nets($r=10$)            & 33.30          & 0.10    & 4.8 $\times$ \\
\textbf{PSTRN-M(Ours)} & \textbf{33.25} & 0.10    & 5.2 $\times$ \\
\hline
TR-Nets($r=15$)$^{ri}$       & 32.73         & 0.227    & 2.1 $\times$ \\
\textbf{PSTRN-S(Ours)} & \textbf{31.95} & 0.226    & 2.1 $\times$ \\
\hline
\end{tabular*}
\end{table}

\begin{table}[t]
\renewcommand\arraystretch{1.2}
\centering
\caption{Results of TR decomposition for LSTM on HMDB51.}
\label{HMDB51_A}
\begin{tabular*}{8.4cm}{lccc}
\hline
\textbf{Model} & \textbf{Error(\%)} & \textbf{Params(M)} & \textbf{CR}\\
\hline
Original               & 51.85          & 17         & 1              \\
TR-LSTM($r=40$)$^{ri}$            & 42.98         & 0.46          & 36.41$\times$  \\
TR-LSTM($r=60$)$^{ri}$            & 41.95         & 1.04          & 16.2$\times$  \\
\hline
\textbf{PSTRN-M(Ours)} & 42.98 & 0.15          & 114.79$\times$  \\
\textbf{PSTRN-S(Ours)} & \textbf{40.85} & 0.41         & 41.37$\times$  \\
\hline
\end{tabular*}
\end{table}

\end{document}